\documentclass[lettersize,journal]{IEEEtran}
\usepackage{amsmath,amsfonts}
\usepackage{algorithmic}
\usepackage{algorithm}
\usepackage{array}
\usepackage[caption=false,font=normalsize,labelfont=sf,textfont=sf]{subfig}
\usepackage{textcomp}
\usepackage{stfloats}
\usepackage{url}
\usepackage{verbatim}
\usepackage{graphicx}
\usepackage{cite}
\usepackage{multirow}
\usepackage{booktabs}
\usepackage{xcolor}

\begin{document}

\title{UniFGVC: Universal Training-Free Few-Shot Fine-Grained Visual
Classification via Attribute-Aware Multimodal Retrieval}

\author{Haiyun Guo, Hongyu Guo, Xiangzhao Hao, Jiarui Guo, Jinqiao Wang, Tat-Seng Chua

\thanks{Haiyun Guo, Jinqiao Wang are with the Foundation Modal Research Center, Institute of Automation, Chinese Academy of Sciences, Beijing 100190, China(e-mail: haiyun.guo@nlpr.ia.ac.cn; jqwang@nlpr.ia.ac.cn).}
\thanks{Hongyu Guo is with the School of Traffic and Transportation, Beijing Jiaotong University, Beijing 100044, China(e-mail:23120800@bjtu.edu.cn).}
\thanks{Xiangzhao Hao is with the School of Artificial Intelligence, Shanghai Jiao Tong University, Shanghai 200030, China(e-mail:xiangzhao.hao@nlpr.ia.ac.cn).}
\thanks{Jiarui Guo is with the University of Hong Kong, Hongkong 999077, China(e-mail:guojiarui@bupt.edu.cn).}
\thanks{Tat-Seng Chua is with the Department of Computer Science, NUS School of Computing, Singapore 117417(e-mail:chuats@comp.nus.edu.sg).}
\thanks{Corresponding author: Jinqiao Wang.}
}




\maketitle

\begin{abstract}

Few-shot fine-grained visual classification (FGVC) aims to discriminate between subtly different categories with very limited data. While  Multimodal Large Language Models (MLLM) exhibit strong capabilities in both fine-grained visual understanding and open-world category recognition, directly fine-tuning them on few-shot FGVC tasks typically leads to overfitting on sparse samples, which erodes their valuable open-world knowledge and compromises generalization.
To fundamentally overcome this limitation, we present UniFGVC, a universal training-free framework that shifts the paradigm from parameter fine-tuning to training-free retrieval-based inference. UniFGVC reframes few-shot FGVC as a multimodal retrieval problem using predefined category templates.
Its core is the Category-Discriminative Visual Captioner (CDV-Captioner), an attribute-aware module designed to leverage MLLM for generating structured textual descriptions that explicitly pinpoint discriminative visual attributes between similar classes. By integrating chain-of-thought reasoning and visual references, CDV-Captioner mitigates hallucination and maximizes the discriminative power of textual captions, thereby converting images into attribute-enriched image-description pairs.
Subsequently, utilizing arbitrary pre-trained visual and textual encoders in a plug-and-play manner, UniFGVC extracts features from both modalities and computes their similarity with all category templates for recognition. This training-free design universally accommodates various encoders and MLLM while fully preserving their original capabilities, ensuring robust generalization across diverse few-shot FGVC scenarios without any task-specific modifications. 
Extensive experiments on 12 FGVC datasets demonstrate UniFGVC's universal compatibility and consistent superiority. It not only outperforms existing few-shot CLIP-based methods but also surpasses several fully-supervised MLLM-based approaches, confirming the general advantage of the proposed retrieval paradigm.
\end{abstract}


\section{Introduction}

\IEEEPARstart{F}{ine-grained} visual classification (FGVC) aims to discriminate between visually similar subcategories within a broader superordinate category (e.g., bird species or car models). It is critical in fields like biodiversity monitoring\cite{wang2023learning,he2025analyzing}, intelligent retail\cite{peng2020rp2k}, and smart transportation\cite{tafazzoli2017large}. In these real-world scenarios, FGVC systems must overcome large intra-class variation, small inter-class differences, and cross-domain shifts, which traditionally demand extensive expert annotation. To alleviate this burden and enable rapid adaptation to novel categories and domains, few-shot FGVC has been introduced \cite{huang2020low,zhang2021tip,peng2023sgva,zhou2022conditional,wei2024enhancing}.  
The core challenge of this task lies in learning highly discriminative representations and achieving robust generalization from only a few labeled examples.


\begin{figure}[t]
  \centering
  \includegraphics[width=1.0\columnwidth]{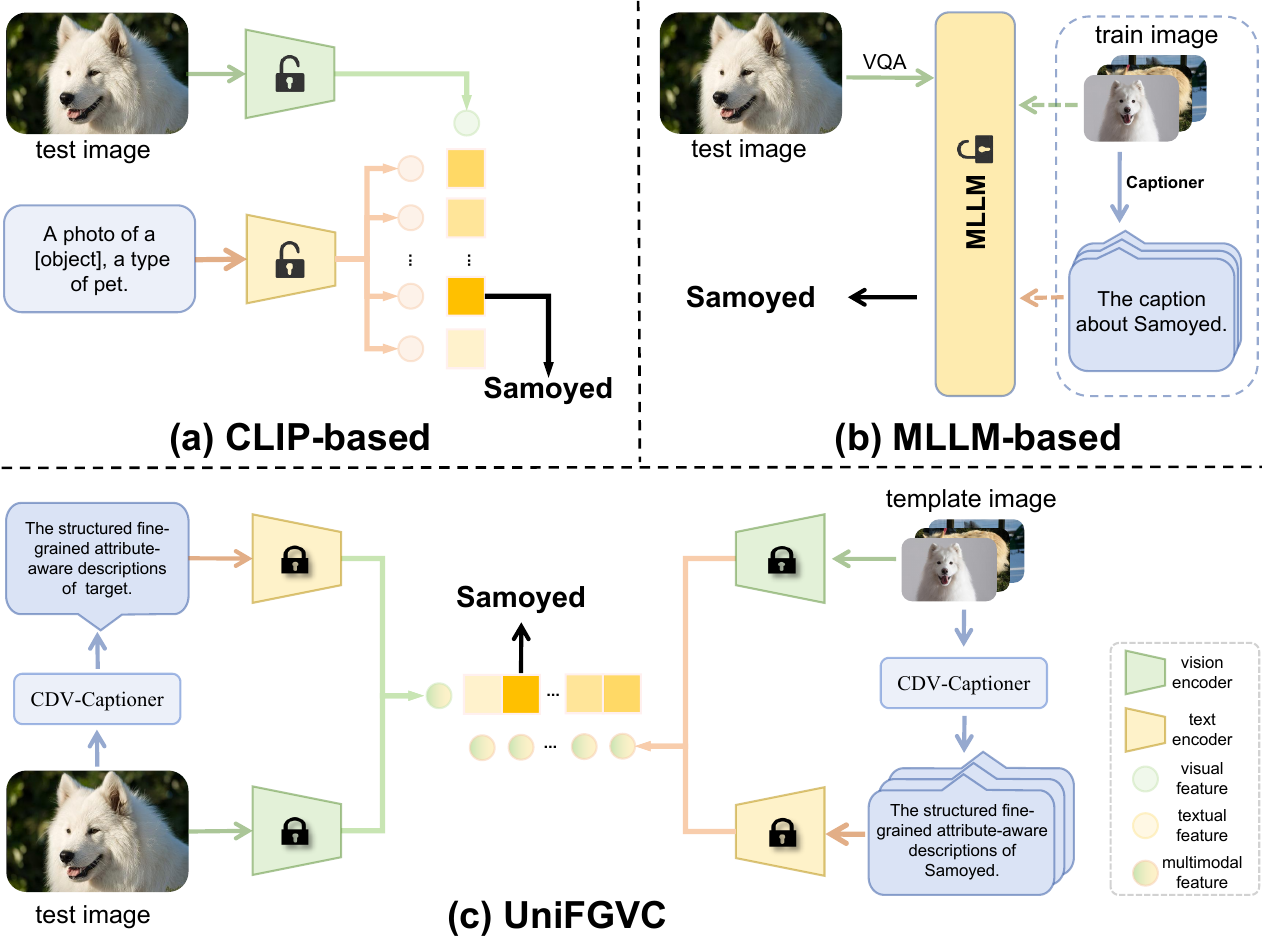}
    \caption{Overview of different few-shot FGVC paradigms. (a) CLIP-based methods rely on fine-tuning to achieve fine-grained discrimination but show limited cross-domain generalization. (b) MLLM-based methods leverage generated image captions in model training to enhance fine-grained recognition, but the captions are often generic or hallucinated. (c) Our proposed UniFGVC shifts the paradigm from parameter fine-tuning to training-free retrieval-based inference, reframing the task as a multimodal retrieval problem with predefined category templates. Central to UniFGVC is the CDV-Captioner, which leverages chain-of-thought reasoning and visual references to mitigate hallucination and produce structured attribute-enriched textual descriptions. }
  \label{fig:introduction}
\end{figure}


Recent vision-language models (VLM), exemplified by CLIP\cite{radford2021learning}, and Multimodal Large Language Models (MLLM) offer a promising foundation for FGVC due to their strong open-world recognition and fine-grained visual understanding capabilities. However, adapting them to few-shot FGVC remains challenging, as prevailing methods rely on training-dependent paradigms with fundamental limitations. 
As shown in Figure~\ref{fig:introduction}(a), one mainstream strategy performs lightweight fine-tuning of VLM through learnable prompts\cite{zhou2022learning}, adapters\cite{zhang2021tip}, or cache prototypes\cite{schick2020exploiting}. Although parameter-efficient, this paradigm inevitably leads to overfitting on few-shot data\cite{xu2023fine,xu2023improving}, which critically erodes the model's valuable pre-trained, open-world knowledge and compromises generalization to novel categories and domains\cite{li2021supervision}.
A parallel line of work, illustrated in Figure~\ref{fig:introduction}(b), fine-tunes MLLM\cite{liu2024democratizing} for FGVC, typically using textual descriptions to augment training data. Beyond sharing the overfitting risk, these approaches face another challenge. The generated descriptions are prone to hallucination and lack discriminative attributes (e.g., the coat texture differences between Golden Retrievers and Labradors)   necessary for fine-grained separation. Consequently, existing methods are constrained by the need for error-prone training on scarce data. This critical gap motivates our investigation into a fundamentally different, training-free paradigm that can fully harness the rich pre-trained knowledge within VLM/MLLM for universally generalizable few-shot FGVC.

To bridge this gap, we propose UniFGVC, a universal training-free framework for few-shot FGVC that fundamentally shifts the paradigm. Its innovation lies in a threefold decoupling. Firstly, it decouples task adaptation from parameter updates via a retrieval paradigm to preserve pre-trained knowledge and eliminate overfitting. Secondly, it decouples discriminative knowledge from the model's implicit state into explicit, structured descriptions via our novel Category-Discriminative Visual Captioner (CDV-Captioner). Thirdly, it decouples the framework from specific models via a modular design for universal plug-and-play compatibility. Consequently, UniFGVC inherently ensures robust generalization, fine-grained discriminability, and broad applicability. 

At the core of UniFGVC is our CDV-Captioner, an attribute-aware module designed to tackle the dual issues of hallucination and non-discriminative descriptions. It leverages the inherent capabilities of MLLM through a reference-guided Chain-of-Thought (CoT) process.  As shown in Figure~\ref{fig:method}, for a target image, we first retrieve a few visually similar reference images from different yet highly similar categories. The MLLM is then progressively guided to: 1) Analyze the references to establish fine-grained distinction criteria and identify key discriminative regions in the target image; 2) Describe the attribute characteristics of each region; 3) Summarize these region-wise attributes into a compact, structured textual description. This process directly elicits the discriminative power of MLLM as well as mitigate hallucinations. Instead of generating long captions with exhaustive details, CDV-Captioner exclusively describes the most discriminative attributes and structures them compactly to enhance both efficiency and accuracy of retrieval. 

Based on this, we construct a training-free multimodal retrieval pipeline. All training images are converted into attribute-enriched image-description pairs via CDV-Captioner. Then, arbitrary pre-trained vision and text encoders can be used in a plug-and-play manner to extract and fuse features into a multimodal template gallery. During inference, a target image follows the same process, and its multimodal features are matched against all templates via similarity calculation for category prediction.

Notably, while the inference involves multi-step reasoning, the fully training-free nature of UniFGVC offers practical advantages in total cost and operational flexibility. Compared to methods requiring per-task fine-tuning, UniFGVC incurs a one-time cost to build the template gallery, which then supports zero-marginal-cost extension to novel categories. The inference relies on efficient forward passes and feature matching, and all components including MLLM and encoders can leverage lightweight or optimized variants to meet specific latency requirements, as discussed in Section~\ref{sec:eff}.

Our contributions are summarized as follows:
\begin{itemize}
\item We propose UniFGVC, a universal training-free framework that fundamentally shifts the paradigm for few-shot FGVC from parameter fine-tuning to multimodal retrieval. This design decouples task adaptation from parameter updates through retrieval, and knowledge elicitation from model internals through generating explicit discriminative descriptions. Thereby, it preserves pre-trained knowledge and inherently ensures robust generalization.

\item We propose CDV-Captioner, an attribute-aware module that tackles hallucination and non-discriminative descriptions by eliciting structured, attribute-enriched captions from MLLM through a reference-guided chain-of-thought reasoning process.

\item We extensively evaluate UniFGVC on 12 datasets. On average, it outperforms state-of-the-art few-shot FGVC methods by 4.62\%, with a notable gain of 12.31\% on ImageNet, and even surpasses several fully supervised MLLM-based models. Ablation studies with different MLLM and encoders further validate the broad adaptability and effectiveness of our framework.



\end{itemize}

\section{Related Work}

\subsection{Few-Shot Fine-Grained Classification}

Recent few-shot fine-grained visual classification (FGVC) methods often rely on powerful pre-trained visual-language models (VLM) such as CLIP\cite{radford2021learning}, which establish robust image-text alignment through joint visual-linguistic representation learning\cite{radford2021learning,an2023unicom,koukounas2024jina}. These approaches typically adapt contrastive VLM with a small number of labeled samples to align vision and text encoders within a unified embedding space. Building on large-scale image-text pretraining, methods such as CoOp\cite{zhou2022learning}, Tip-Adapter\cite{zhang2021tip}, and related approaches enhance FGVC performance by introducing lightweight adaptation mechanisms that refine the alignment between visual and textual features, thus improving inter-class discrimination while mitigating overfitting from limited data\cite{qu2025proapo,gao2022pyramidclip}.

Capitalizing on the powerful zero-shot capabilities of VLM, Tip-Adapter\cite{zhang2021tip} further improved few-shot adaptation by introducing a key-value cache model that facilitates efficient training and enables rapid convergence. Methods like CoOp\cite{zhou2022learning} automated prompt optimization to enhance performance with minimal labeled data, while CoCoOp\cite{zhou2022conditional} incorporated automated prompts to improve textual feature representations, thereby alleviating the limitations of purely vision-based embeddings. SgVA-CLIP\cite{peng2023sgva} learned adapted visual features that complement the cross-modal space, thereby capturing fine-grained discriminative details that were overlooked by purely text-aligned representations. CaFo\cite{zhang2023prompt} integrated multiple foundation models in a cascaded manner to enhance few-shot learning. T-IDEA\cite{ye2025idea} enhanced few-shot image classification by leveraging CLIP's dual encoders to compute multimodal similarities between test images and image-text pairs from a support set. In addition, ProKeR\cite{bendou2025proker} employed kernel-based regularization for VLM adaptation.


Existing CLIP-based few-shot FGVC methods often suffer from severe overfitting when fine-tuned on limited samples, and they also fail to fully exploit the rich open-world priors embedded in the underlying vision–language models. Moreover, these methods lack extensibility, as they require task-specific retraining to accommodate new categories, thereby incurring non-negligible marginal costs for each addition. In contrast, our training-free framework effectively leverages fine-grained inter-category semantics and inherent world knowledge, while supporting zero-marginal-cost extension to novel categories, resulting in more robust and accurate classification.

\begin{figure*}[t]
  \centering
  \includegraphics[width=1.0\textwidth]{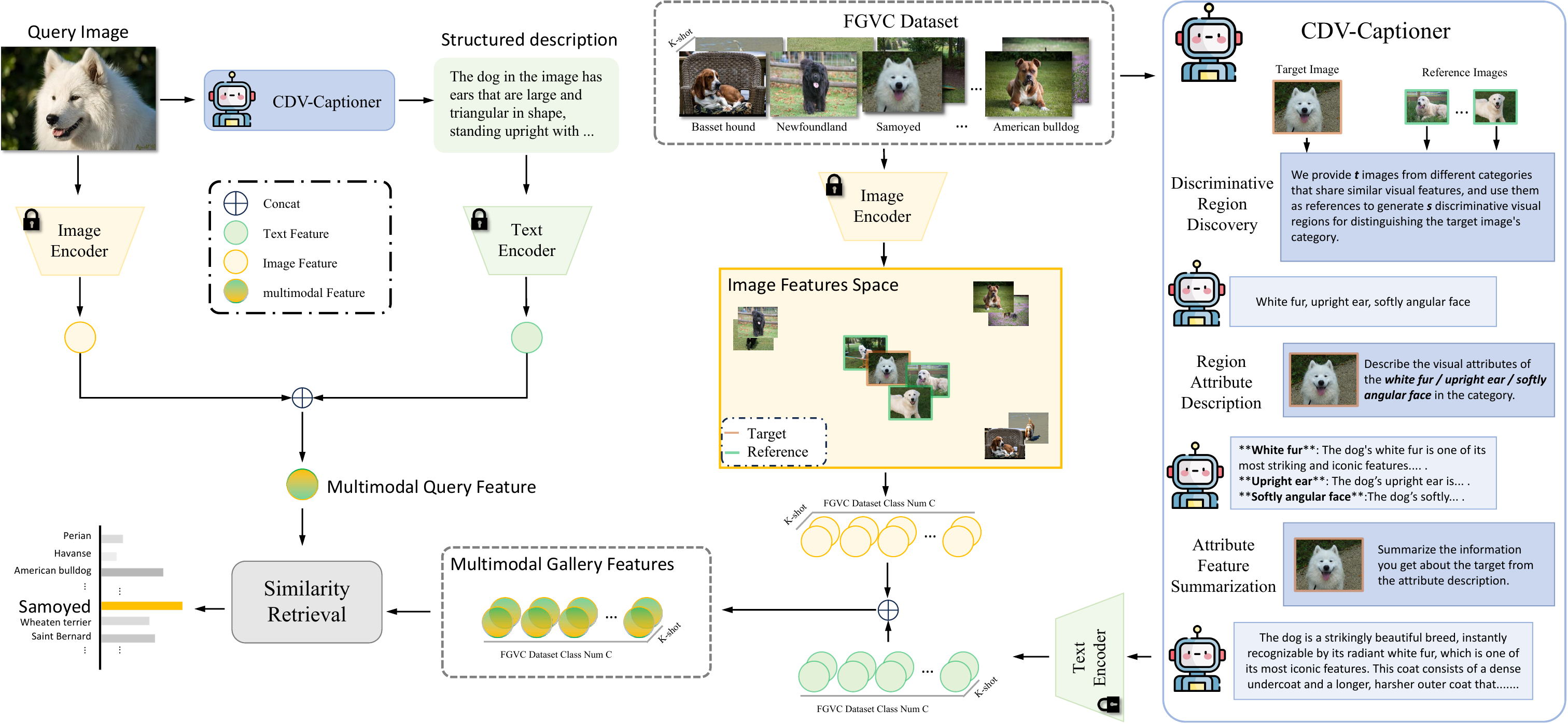}
  \caption{An overview of the proposed UniFGVC. UniFGVC is a universal, training-free framework for few-shot fine-grained visual classification, which reformulates the task as a multimodal retrieval problem using structured attribute-aware representations. The CDV-Captioner progressively prompts the MLLM to output the structured fine-grained attribute-aware feature description of the target image, by integrating the category-related linguistic priors inherent in the MLLM and visual priors derived from reference images.}
  \label{fig:method}
\end{figure*}

\subsection{Multimodal Large Models for Fine-Grained Classification}

Recent advances, exemplified by Qwen2.5-VL\cite{bai2023qwen}, InternVL\cite{chen2024far} and GPT-4o\cite{achiam2023gpt}, have demonstrated remarkable effectiveness of MLLM in translating intricate visual patterns into textual descriptions that capture subtle distinctions in texture, morphology, and shape configurations. By bridging high-level visual concepts with linguistically grounded expressions, MLLM have provide enriched multimodal representations that enable attribute-aware FGVC, significantly improving classification accuracy over conventional vision-only approaches\cite{menon2022visual,liu2024democratizing,chen2024far}. Existing MLLM-based FGVC methods often adopt a parameter fine-tuning paradigm to leverage the powerful representational capacity of MLLM.

FineR\cite{liu2024democratizing} employed large language models to translate visual attributes into textual descriptions, enabling category identification without expert-defined labels. CascadeVLM\cite{wei2024enhancing} utilized MLLM for FGVC by prompting them with reranked candidate classes but lacked explicit modeling of fine-grained distinctions. Finedefics\cite{he2025analyzing} improved recognition performance by incorporating object attribute descriptions during training and using contrastive learning to align relationships between visual objects, attributes, and categories. More recently, an analysis of Idefics\cite{laurenccon2024matters} revealed that training data quality, task-specific alignment, and the balance between visual and linguistic representations were critical for achieving robust fine-grained performance, providing valuable design insights beyond ad hoc model improvements. 


Existing MLLM-based fine-grained classification methods either use image captions as supervision for training or rely on MLLM reasoning to predict category labels. However, MLLM often fail to capture and reason about the most discriminative attributes required for fine-grained category differentiation. Without task-specific guidance, the generated captions tend to be generic and are prone to hallucination, introducing noisy or irrelevant information. In contrast, our method constructs structured, attribute-aware textual descriptions tailored to the target task and leverages target-specific semantic priors during inference, enabling more precise and robust fine-grained discrimination without additional training.

\subsection{Attribute-Based Fine-Grained Classification}
Attribute-based learning has long been a foundational paradigm for fine-grained recognition. Early studies showed that subtle category differences can be better captured through localized and semantically meaningful attributes rather than holistic appearance alone. This idea was explored through localized attribute discovery\cite{duan2012discovering}, attribute-guided attention\cite{liu2017localizing}, and dense attribute-based recognition\cite{huynh2020fine}, where attributes serve as explicit intermediate cues for identifying discriminative regions and supporting category decisions. Later work extended this direction from fixed attribute vocabularies to richer natural-language descriptions, including fine-grained visual descriptions\cite{reed2016learning} and compositional attribute phrases\cite{su2017reasoning}, highlighting that class distinctions can be expressed more flexibly in language. In the foundation-model era, this line has further evolved into description-based and attribute-guided recognition with VLMs, where language models provide candidate attributes\cite{menon2022visual} or prompts\cite{tian2024argue} for classification.

Our work is closely related to this tradition, since it is also built on the principle that fine-grained recognition should be grounded in explicit discriminative attributes. However, UniFGVC differs from prior attribute-based methods in two key aspects. First, unlike classical approaches that rely on predefined attribute vocabularies, manual attribute annotations, or dedicated attribute predictors, our method elicits open-vocabulary, reference-conditioned attribute descriptions from an MLLM. Second, these attributes are not used as auxiliary supervision for parameter learning. Instead, they are converted into structured textual representations for training-free multimodal retrieval. Therefore, our method can be viewed as bringing the classical attribute-based intuition into a training-free MLLM framework for few-shot FGVC.

\section{Method}
This section introduces the UniFGVC framework. We first present the Category-Discriminative Visual Captioner (CDV-Captioner), a structured description module that employs multimodal chain-of-thought prompting to guide MLLM in adaptively identifying the key discriminative regions within target images. These regions are then translated into fine-grained structured textual descriptions that capture subtle visual distinctions essential for FGVC. Based on these descriptions, each image is converted into an image-text pair, forming a rich multimodal representation for downstream retrieval. We then extract hybrid visual-linguistic features using any off-the-shelf pre-trained encoders and perform fine-grained category recognition by computing feature similarity against a multimodal category template database.

\subsection{Category-Discriminative Visual Captioner}
While prior work has demonstrated the efficacy of language in semantic modeling\cite{he2025analyzing}, image-text alignment relying exclusively on category names exhibits inherent limitations due to insufficient discriminative signals. In contrast, region-based semantic representations offer richer structural information for fine-grained differentiation. Building on recent advances in multimodal learning, we propose a CDV-Captioner, which adopts a chain-of-thought prompting strategy to progressively guide MLLM in identifying and articulating the most discriminative regions through comparative reasoning with reference samples. It then converts these region-level insights into structured textual descriptions that encode semantically grounded attributes.

Specifically, as illustrated in Figure~\ref{fig:method}, the CDV-Captioner operates through four coordinated stages: 1) \textbf{Reference Sample Selection.} For each target image, we retrieve a set of visually similar exemplars from a shared feature space constructed from K-shot training samples. 2) \textbf{Discriminative Region Discovery.} Through comparative analysis with reference samples, the MLLM progressively localize the most discriminative visual regions in the target image, those that most effectively distinguish it from similar categories. 3) \textbf{Region Attribute Description.} The MLLM generate detailed attribute descriptions for each identified region, capturing fine-grained characteristics. These descriptions explicitly encode categorical distinctions to differentiate between highly similar classes. 4) \textbf{Attribute Feature Summarization.} The MLLM processing stage consolidates multiple region-specific descriptions into a unified structured textual description. These descriptions integrate comprehensive attribute information essential for discriminating between fine-grained categories.

\textbf{Reference Sample Selection.} To effectively guide the generation of fine-grained descriptions, the CDV-Captioner adopts a reference-guided exemplar selection strategy. Specifically, given a \textit{K}-shot dataset with \textit{C} classes and \textit{K} support images per class, let \begin{math}
    \ f_{c,j}
\end{math} denote the feature of the \textit{j}-th support image in class \textit{c}, extracted by a pre-trained image encoder. We define the prototype of class \textit{c} as: 
\begin{equation}
    \ p_c = \frac{1}{K}\sum_{j=1}^{k}{f_{c,j}}
\end{equation}
For a target image with feature \begin{math}
    \ f_t
\end{math} extracted by the same encoder, we first retrieve the top-\textit{t} nearest class prototypes from the \textit{K}-shot support set:
\begin{equation}
  \ I_{ref} = NN(f_t,\{p_c\}^C_{c=1})
\end{equation} 
where 
\begin{math}
    NN(\cdot)
\end{math} returns the \textit{t} most similar classes. Then, for each retrieved class \textit{c}, we select one representative exemplar as the support image closest to its class prototype. The final reference set is   
\begin{math}
    \ I_{ref} = \left \{I^1_{ref},...,I^t_{ref}\right \}
\end{math}. 
Thus, the reference images are retrieved exclusively from the few-shot support set and correspond to \(t\) visually similar but distinct categories, providing cross-category visual comparisons for description generation. In the 1-shot setting, each class prototype naturally degenerates to its only support image, i.e., \(p_c=f_{c,1}\), so the reference images are directly selected from the visually most similar support images of different categories.

\textbf{Discriminative Region Discovery.} We utilize the target image along with a set of visually similar reference images as input. Through comparative analysis, the MLLM identify category-discriminative structural regions, such as white fur, upright ears, and a softly angular face, that most effectively capture inter-class differences. These regional cues enable precise differentiation among fine-grained categories. Specifically, we used MLLM: ["\{IMAGERY\} We provide \{t\} images from different categories that share similar visual features, and use them as references to generate \{s\} discriminative visual regions for distinguishing the target image's category."]. Formally, MLLM takes a target image
\begin{math}
  \ I_{t} 
\end{math} and \textit{t} reference images 
\begin{math}
  \ I_{ref} 
\end{math} as input. It outputs a set of discriminative region names:
\begin{equation}
  \ N^{t} = W_{\theta }(P^{dis} (I_{t}, I_{ref}))
\end{equation} 
where 
\begin{math}
  \ N^{t} = \left \{ N_1^{t},...,N_i^{t},..., N_s^{t}\right \}
\end{math} denotes the \textit{s} regions for target image 
\begin{math}
  \ I_{t} 
\end{math}, 
\begin{math}
  \ W_{\theta} 
\end{math} is the MLLM, and 
\begin{math}
  \ P^{dis}
\end{math} is the discriminative region discovery prompt.

\textbf{Region Attribute Description.} With the discovered region names 
\begin{math}
  \ N^{t}
\end{math}, we leverage the strong capability of MLLM to recognize generic visual attributes and produce multiple attribute-specific descriptions for each region. For example, when processing a set of regions
\begin{math}
  \ N^{t}
\end{math}, MLLM generate concise descriptions of the characteristics of the target's regions, providing a more tractable and informative representation compared to directly discriminating among fine-grained subordinate categories. Specifically, we used MLLM: ["\{IMAGE\} Describe the visual attributes of the \{
\begin{math}
    N^{t}
\end{math}
\} in the category."]. Formally, MLLM takes the target image 
\begin{math}
  \ I_{t} 
\end{math} and the region names
\begin{math}
  \ N^{t}
\end{math} as input, and outputs a visual attributes description:
\begin{equation}
  \ V = W_{\theta }(P^{reg} (I_{t},N^{t}))
\end{equation} 
where 
\begin{math}
  \ V = \left \{ V_1,...,V_i,..., V_s\right \}
\end{math} contains the attribute-level descriptions for each region, and
\begin{math}
  \ P^{reg}
\end{math} is the region attribute description prompt.

\textbf{Attribute Feature Summarization.} Upon acquiring the structured attribute image-description pairs, we used MLLM: ["\{IMAGE\} Summarize the information you get about the target from the attribute description."]. The summarized description is rich in fine-grained attribute information, enabling more precise characterization of subordinate-level semantic features for effective inter-category discrimination. Formally, given the region names 
\begin{math}
  \ N^{t}
\end{math} and attribute descriptions \begin{math}
  \ V
\end{math}, the MLLM outputs a summarized attribute description \begin{math}
    \ A_{i}
\end{math} for the target image 
\begin{math}
  \ I_{t}
\end{math}:
\begin{equation}
  \ A_{i} = W_{\theta }(P^{sum} (I_{t},V,N^{t}))
\end{equation} 
where 
\begin{math}
  \ A_{i}
\end{math} is the fine-grained textual description, and 
\begin{math}
  \ P^{sum}
\end{math} is the summarization prompt. The CDV-Captioner transforms conventional image-category pairs into enriched image-description-category tuples by generating discriminative structured textual descriptions. These descriptions serve as semantic bridges that explicitly connect visual instances with their fine-grained categorical labels through attribute-level feature representations.

\subsection{FGVC via Multimodal Retrieval}
Our proposed UniFGVC framework reformulates the few-shot FGVC task as a multimodal fine-grained category retrieval problem with predefined category templates. The entire framework is completely training-free, which requires neither parameter updates nor task-specific prompt tuning, dataset-specific calibration, or additional optimization. In particular, UniFGVC uses a fixed prompt template together with a generic reference selection strategy across all datasets. Under this setting, it constructs a high-precision retrieval database from minimal support samples by generating attribute-rich representations, while preserving the generalization capacity of foundation models. This distinguishes UniFGVC from methods that avoid weight updates but still rely on task-specific prompt engineering or dataset-dependent calibration.

The training-free framework implements a hybrid retrieval paradigm operates through two core components: 1) \textbf{Multimodal Category Template Database Construction.} The CDV-Captioner automatically generates structured textual descriptions to populate the retrieval database, where image-description pairs are encoded as aligned multimodal embeddings through feature fusion. This process preserves fine-grained attribute relationships critical for category discrimination. 2) \textbf{FGVC via Multimodal Retrieval.} UniFGVC performs category prediction via nearest-neighbor retrieval in the multimodal database space. Similarity-based matching leverages both visual-semantic alignment and attribute-level discriminative signals to identify optimal category assignments without requiring model fine-tuning.

\textbf{Multimodal Category Template Database Construction.} Given a \textit{K}-shot dataset spanning \textit{C} categories, we apply the CDV-Captioner to generate structured textual descriptions, denoted as 
\begin{math}
  \ T_{i}
\end{math}, for each training image. These image-description pairs form the foundation of our retrieval database, encapsulating the few-shot knowledge across \textit{C} classes. For each image-description pair, we use a pre-trained image encoder to extract visual features, and a pre-trained text encoder to extract textual features from the description. The two modalities are fused to form a unified multimodal representation for retrieval:
\begin{equation}
  \ F_{i} = Fusion(I_E(I_{i}),T_E(T_{i}))
\end{equation}
where 
\begin{math}
    \ I_{E}
\end{math} is the image encoder, 
\begin{math}
    \ T_{E}
\end{math} is the text encoder, and \textit{Fusion} is a multimodal feature integration strategy, where we concatenate the visual and textual feature vectors to construct the joint representation. 

For all \textit{CK} gallery samples, we denote their fused features as \begin{math}
    \ F_{gallery} \in \mathbb{R} ^{CK\times N} 
\end{math}, where N is the dimension of the fused feature. The corresponding labels are stored in a vector\begin{math}
    \ L_{label} \in \mathbb{R} ^{CK}
\end{math}. To enhance the generalizability and precision of gallery representations, instead of generating textual descriptions for each individual image, we aggregate information from all \textit{K} images within the same class to construct a unified class-level textual description. This ensures that the gallery features capture category-specific semantics while avoiding instance-level noise. For the key-value cache, the 
\begin{math}
    \ F_{gallery}
\end{math} is treated as the keys, and \begin{math}
    \ L_{label}
\end{math} as the values. In contrast, query representations are constructed for each target image, enabling effective retrieval by comparing target-specific queries with generalized class-level gallery representations. Thus, the retrieval database encapsulates the task-specific knowledge derived from the few-shot training set.

\textbf{FGVC via Multimodal Retrieval.} After constructing the retrieval database, feature matching can be achieved through simple matrix operations that compute multimodal similarity scores. During inference, the target image is processed through the CDV-Captioner to generate structured textual descriptions. These descriptions, along with the original image, are then encoded into joint visual-textual features
\begin{math}
    \ F_{query} \in \mathbb{R} ^{1\times N}
\end{math} using their respective pre-trained encoders. The fused query features are then used to perform similarity-based retrieval against the pre-constructed database. The similarity scores between the query and all gallery samples are computed as 
\begin{equation}
  \ R = NN(F_{query},F^T_{gallery})
\end{equation}
where 
\begin{math}
    \ R \in \mathbb{R}^{1\times CK}
\end{math} is the similarity vector between the query feature and all gallery features. The operator 
\begin{math}
    NN(\cdot)
\end{math} represents a nearest-neighbor retrieval function, which computes the similarity between the query feature
\begin{math}
    \ F_{query}
\end{math} and all few-shot gallery features
\begin{math}
    \ F^T_{gallery}
\end{math}. The final classification is obtained by retrieving the label of the gallery sample with the highest similarity score to the query.

\begin{table*}[t]
  \centering
  \caption{Comparative performance(\%) of different few-shot FGVC methods across 1/2/4/8/16-shot settings on 11 fine-grained classification datasets. \textbf{Bold} and \underline{underline}  indicates the best and second best performance.}
  \label{tab:results}
  \renewcommand{\arraystretch}{0.9}
  \begin{tabular}{lcccccccccccccc}
  \toprule
    Methods & Venue & Setting & Img. & Cal. & DTD & Eur. & Air. & Flo. & Food & Pets & Cars & SUN & UCF & avg.  \\\midrule
    CLIP & ICML2021 & Zero-shot & 58.18 & 86.29 & 42.32 & 37.56 & 17.28 & 66.14 & 77.31 & 85.77 & 55.61 & 58.52 & 61.46 & 58.77\\\toprule
     CoOp & IJCV2022 & \multirow{9}{*}{\centering 1-shot} & 57.15 & 87.53 & 44.39 & 50.63 & 9.64 & 68.12 & 74.32 & 85.89 & 55.59 & 60.29 & 61.92 & 59.59 \\
     Tip-Adapter & ECCV2022 &  & 60.88 & 88.80 & 50.49 & 50.34 & 19.01 & 81.17 & 76.22 & 86.04 & 56.78 & 61.23 & 66.19 & 63.38\\
     SgVA-CLIP & TMM2023 &  & 61.11 & 89.67 & 49.97 & 58.18 & 22.34 & 76.17 & 78.16 & 88.05 & 58.17 & 64.81 & 64.43 & 64.64\\
     CaFo & CVPR2023 &  & 63.81 & \underline{91.82} & \underline{53.14} & \textbf{68.24} & 25.01 & 81.29 & 77.98 & 89.21 & 62.12 & 64.87 & 68.71 & 67.84\\
     GDA & ICLR2024 &  & 61.07 & 87.36 & 47.27 & 58.07 & 17.62 & 72.12 & 77.41 & 85.51 & 56.87 & 59.97 & 62.81 & 62.37\\
     TransCLIP & NeurIPS2024 &  & \underline{64.32} & \textbf{92.17} & 52.31 & 63.51 & 22.68 & 83.23 & \textbf{82.24} & \textbf{90.10} & \underline{63.68} & \textbf{68.88} & \textbf{69.51} & 68.44\\
     T-IDEA & PR2025 &  & 61.32 & 89.78 & 51.17 & 64.32 & 21.76 & 82.07 & 78.81 & 87.01 & 58.64 & 62.78 & 66.31 & 65.82\\
     ProKeR & CVPR2025 & & 61.27 & 88.94 & 48.21 & 59.82 & 21.07 & 78.97& 77.61 & 86.08 & 58.61 & 62.17 & 63.97 & 64.25\\
     ProAPO & CVPR2025 &  & 62.58 & 90.68 & \textbf{57.21} & \underline{67.01} & \underline{25.21} & \underline{84.86} & \underline{81.44} & \underline{89.29} & 61.13 & \underline{66.68} & \underline{69.01} & \underline{68.65} \\\midrule
    \textbf{UniFGVC} &  &  & \textbf{70.18} & 91.07 & 50.59 & 56.74 & \textbf{43.05} & \textbf{89.64} & 77.64 & 85.01 & \textbf{81.69} & 57.20 & 67.77 & \textbf{70.06}\\\toprule
    CoOp & IJCV2022 & \multirow{9}{*}{\centering 2-shot} & 57.81 & 87.93 & 45.15 & 61.50 & 18.68 & 77.51 & 72.49 & 82.64 & 58.28 & 59.48 & 64.09 & 62.32\\
    Tip-Adapter & ECCV2022 &  & 61.57 & 89.61 & 55.32 & 64.76 & 21.76 & 85.40 & 77.05 & 86.06 & 61.13 & 63.19 & 68.99 & 66.80\\
    SgVA-CLIP & TMM2023 &  & 62.74 & 90.63 & 53.24 & 65.79 & 23.16 & 82.03 & 78.07 & 88.28 & 61.94 & 66.23 & 67.52 & 67.24\\
    CaFo & CVPR2023 &  & 64.34 & \textbf{92.41} & \underline{57.01} & \underline{73.22} & \underline{26.71} & 84.98 & 78.07 & 89.02 & 63.17 & 66.61 & 70.73 & 69.66\\
    GDA & ICLR2024 &  & 61.21 & 88.64 & 52.33 & 67.32 & 22.56 & 83.07 & 77.89 & 86.64 & 59.16 & 63.76 & 66.40 & 66.27\\
    TransCLIP & NeurIPS2024 &  & \underline{64.77} & \underline{92.08} & 52.86 & \textbf{76.85} & 22.88 & 85.42 & \underline{81.23} & \textbf{89.87} & 63.84 & \textbf{69.52} & \underline{71.03} & 70.03\\
    T-IDEA & PR2025 &  & 61.78 & 90.31 & 56.01 & 68.02 & 24.43 & 84.87 & 77.91 & 87.38 & 62.07 & 64.87 & 68.12 & 67.80\\
    ProKeR & CVPR2025 &  & 61.42 & 89.41 & 53.04 & 70.21 & 23.66 & 85.03 & 77.82 & 87.34 & 62.09 & 63.81 & 68.04 & 67.44\\
    ProAPO & CVPR2025 &  & 63.44 & 91.34 & \textbf{58.64} & 72.37 & 26.09 & \underline{90.04} & \textbf{81.52} & \underline{89.61} & \underline{64.19} & \underline{67.82} & \textbf{71.29} & \underline{70.58} \\\midrule
    \textbf{UniFGVC} &  &  & \textbf{74.42} & 90.83 & 54.63 & 67.24 & \textbf{47.97} & \textbf{92.10} & 76.71 & 86.56 & \textbf{86.26} & 62.62 & 69.14 & \textbf{73.69} \\\toprule
    CoOp & IJCV2022 & \multirow{9}{*}{\centering 4-shot} & 59.99 & 89.55 & 53.49 & 70.18 & 21.87 & 86.20 & 73.33 & 86.70 & 62.62 & 63.47 & 67.03 & 66.77\\
     Tip-Adapter & ECCV2022 &  & 62.62 & 90.87 & 60.25 & 69.66 & 26.39 & 89.53 & 77.46 & 86.46 & 64.86 & 65.88 & 72.71 & 69.70\\
     SgVA-CLIP & TMM2023 &  & 63.01 & 90.91 & 58.21 & 75.81 & 26.74 & 87.52 & 78.14 & 88.41 & 65.03 & 67.85 & 71.29 & 70.27\\
     CaFo & CVPR2023 &  & \underline{65.64} & \textbf{93.17} & \underline{61.37} & \textbf{84.63} & \underline{33.07} & 91.76 & 78.25 & \textbf{90.07} & 66.32 & 69.27 & 72.87 & \underline{73.31} \\
     GDA & ICLR2024 &  & 61.73 & 90.76 & 57.87 & 75.63 & 28.31 & 89.82 & 77.81 & 87.29 & 62.72 & 65.17 & 71.21 & 69.85 \\
     TransCLIP & NeurIPS2024 &  & 64.72 & \underline{92.24} & 58.03 & \underline{78.03} & 26.41 & 88.72 & \underline{80.34} & 87.21 & 66.48 & \textbf{70.07} & \textbf{74.31} & 71.51\\
     T-IDEA & PR2025 &  & 62.71 & 91.46 & 59.01 & 75.37 & 26.15 & 90.71 & 78.46 & 87.64 & 65.81 & 67.91 & 71.76 & 70.64 \\
     ProKeR & CVPR2025 &  & 62.03 & 90.87 & 58.67 & 76.07 & 28.44 & 90.81& 78.26 & 87.57 & 65.07 & 66.38 & 72.13 & 70.75\\
     ProAPO & CVPR2025 &  & 63.91 & 92.05 & \textbf{67.02} & 76.02 & 29.96 & \underline{92.31} & \textbf{81.56} & \underline{89.85} & \underline{69.01} & \underline{69.66} & \underline{73.81} & 73.20 \\\midrule
    \textbf{UniFGVC} &  &  & \textbf{78.00} & 91.97 & 58.67 & 75.65 & \textbf{45.45} & \textbf{94.36} & 78.65 & 87.05 & \textbf{88.87} & 68.59 & 72.05 & \textbf{76.31} \\\toprule
    CoOp & IJCV2022 & \multirow{9}{*}{\centering 8-shot} & 61.56 & 90.21 & 59.97 & 76.73 & 26.13 & 91.18 & 71.82 & 85.32 & 68.43 & 65.52 & 71.94 & 69.89\\
     Tip-Adapter & ECCV2022 &  & 64.15 & 91.70 & 62.93 & 79.33 & 30.62 & 91.00 & 77.90 & 88.28 & 69.51 & 69.23 & 74.76 & 72.67 \\
     SgVA-CLIP & TMM2023 &  & 64.23 & 91.79 & 64.05 & 80.29 & 32.51 & 92.50 & 78.67 & 89.16 & 69.88 & 70.26 & 76.28 & 73.60\\
     CaFo & CVPR2023 &  & \underline{66.86} & \textbf{93.85} & 66.81 & \textbf{87.22} & \underline{40.92} & 93.22 & 78.86 & \textbf{90.36} & 70.31 & 70.37 & \underline{78.07} & \underline{76.08} \\
     GDA & ICLR2024 &  & 62.48 & 91.32 & 62.87 & \underline{82.74} & 34.07 & 93.17 & 78.39 & 89.13 & 68.71 & 68.47 & 75.87 & 73.38 \\
     TransCLIP & NeurIPS2024 &  & 65.47 & 92.29 & 61.14 & 78.23 & 28.03 & 89.98 & 80.89 & 89.42 & 70.76 & 71.29 & 76.97 & 73.13\\
     T-IDEA & PR2025 &  & 64.91 & 92.71 & 63.84 & 79.01 & 32.56 & 93.72 & 78.76 & 88.73 & 70.12 & 69.73 & 76.17 & 73.66\\
     ProKeR & CVPR2025 &  & 63.11 & 92.02 & 65.01 & 82.03 & 34.08 & 93.81 & 78.68 & 88.36 & 70.91 & 68.55 & 76.37 & 73.90\\
     ProAPO & CVPR2025 &  & 64.87 & \underline{93.26} & \underline{67.21} & 81.23 & 34.21 & \underline{95.06} & \underline{82.13} & 89.99 & \underline{71.94} & \underline{71.85} & 77.52 & 75.39 \\\midrule
    \textbf{UniFGVC} &  &  & \textbf{80.24} & 92.01 & \textbf{68.99} & 81.69 & \textbf{58.09} & \textbf{96.05} & \textbf{82.22} & \underline{90.32} & \textbf{92.19} & \textbf{73.53} & \textbf{78.73} & \textbf{81.29}\\\toprule
  CoOp & IJCV2022 & \multirow{9}{*}{\centering 16-shot} & 62.95 & 91.83 & 63.58 & 83.53 & 31.26 & 94.51 & 74.67 & 87.01 & 73.36 & 69.26 & 75.71 & 73.42\\
     Tip-Adapter & ECCV2022 &  & 65.44 & 92.63 & 66.94 & 84.94 & 35.86 & 94.23 & 78.11 & 88.18 & 75.75 & 71.00 & 79.03 & 75.65 \\
     SgVA-CLIP & TMM2023 &  & 65.80 & 93.24 & 67.08 & 84.37 & 38.11 & 95.17 & 79.17 & 89.33 & 76.46 & 72.00 & 77.89 & 76.24\\
     CaFo & CVPR2023 &  & \underline{68.79} & \textbf{94.56} & 69.39 & \textbf{88.64} & \underline{48.87} & 95.87 & 79.21 & \textbf{91.54} & 76.37 & 72.41 & 79.66 & \underline{78.66}\\
     GDA & ICLR2024 &  & 63.87 & 92.38 & 67.03 & 87.20 & 41.76 & 96.01 & 79.14 & 88.76 & 75.17 & 70.56 & 77.32 & 76.29 \\
     TransCLIP & NeurIPS2024 &  & 66.64 & \underline{94.42} & 62.73 & 78.87 & 30.09 & 90.88 & 81.14 & 89.52 & 73.21 & 72.58 & 77.24 & 74.30\\
     T-IDEA & PR2025 &  & 66.02 & 93.54 & 67.06 & 84.71 & 38.41 & 95.34 & 79.68 & 90.12 & 76.07 & 71.46 & 78.01 & 76.40\\
     ProKeR & CVPR2025 &  & 64.47 & 93.11 & 67.21 & 86.71 & 41.69 & \underline{96.97} & 79.43 & 89.00 & 76.44 & 72.30 & 78.64 & 76.91\\
     ProAPO & CVPR2025 &  & 66.42 & 93.88 & \underline{70.84} & \underline{87.36} & 40.87 & \textbf{97.01} & \underline{82.26} & 90.66 & \underline{77.23} & \underline{72.69} & \underline{80.09} & 78.12 \\\midrule
    \textbf{UniFGVC} &  &  & \textbf{81.08} & 93.87 & \textbf{73.88} & 84.57 & \textbf{61.06} & 96.34 & \textbf{82.34} & \underline{90.76} & \textbf{94.63} & \textbf{76.55} & \textbf{80.91} & \textbf{83.28}\\\bottomrule
  \end{tabular}
\end{table*}

\section{Experiments}
\subsection{Implementation Details}
\textbf{Datasets.} We evaluate UniFGVC on 12 widely used FGVC datasets: ImageNet\cite{deng2009imagenet}, StanfordCars\cite{krause20133d}, UCF101\cite{ren2015faster},
DTD\cite{cimpoi2014describing}, Caltech101\cite{fei2004learning}, FGVCAircraft\cite{maji2013fine}, Flowers102\cite{nilsback2008automated},  OxfordPets\cite{parkhi2012cats}, Food101\cite{bossard2014food}, SUN397\cite{xiao2010sun}, EuroSAT\cite{helber2019eurosat} and CUBbirds\cite{welinder2010caltech}. For few-shot evaluation, we adopt 1/2/4/8/16-shot configurations and evaluate on the complete official test sets. Notablely, UniFGVC is a training-free approach, yet we compare its performance against existing methods under both few-shot and fully-supervised settings to ensure fair and consistent evaluation across all datasets.


\textbf{Setting.} In our implementation, we adopt Qwen2.5-VL-7B\cite{bai2023qwen} to generate structured textual descriptions for the retrieval database and the test images. For feature encoding and similarity computation, we independently use UniCOM\cite{an2023unicom} as image encoder and Jina-CLIP\cite{koukounas2024jina} as text encoder, without requiring explicit alignment between visual and textual modalities. We set 
\begin{math}
    \ t = 4
\end{math} reference samples per target image and extract
\begin{math}
    \ s = 3
\end{math} structured regions per image to construct discriminative representations.

\begin{table*}[ht]
    \centering
    \caption{Comparative performance (\%) of different fully-supervised MLLM-based models on 5 datasets
    . \textbf{Bold} and \underline{underline} indicates the best and second best performance. \#P denotes parameters count.}
    \label{tab:2}
    \begin{tabular}{lcccccccc}
    \toprule
    Model & Venue & \#P & Flowers102 & OxfordPets & FGVCAircraft & StandfordCars & CUBbirds & Avg. \\\midrule
    LLaVA-Next & CVPR2024 & 7B & 43.91 & 53.72 & 32.49 & 49.48 & 34.88 & 42.90 \\
    FineR & ICLR2024 & 7B & 63.82 & 72.87 & - & 63.46 & 69.54 & - \\
    Idefics2& NeurIPS2024 & 8B & 72.78 & 81.28 & 56.23 & 80.25 & 47.17 & 67.54 \\
    CascadeVLM & EMNLP2024 & 8B & \underline{91.62} & - & \textbf{63.95} & \underline{92.04} & \textbf{80.80} & - \\
    Finedefics& ICLR2025 & 8B & 89.88 & \textbf{92.18} & \underline{63.82} & 84.67 & 57.61 & \underline{77.63} \\\midrule
    \textbf{UniFGVC} & & 7B & \textbf{95.74} & \underline{90.16} & 58.06 & \textbf{94.63} & \underline{77.48} & \textbf{83.21} \\
    \bottomrule
    \end{tabular}
\end{table*}

\subsection{Main Result}

Table~\ref{tab:results} reports the performance of different few-shot FGVC methods across 11 datasets, including ImageNet\cite{deng2009imagenet}, Caltech101\cite{fei2004learning}, DTD\cite{cimpoi2014describing}, EuroSAT\cite{helber2019eurosat}, FGVCAircraft\cite{maji2013fine}, Flowers102\cite{nilsback2008automated}, Food101\cite{bossard2014food}, OxfordPets\cite{parkhi2012cats}, StanfordCars\cite{krause20133d}, SUN397\cite{xiao2010sun} and UCF101\cite{ren2015faster}. Since the majority of few-shot CLIP-based FGVC methods do not report results on CUBird datasets, we present our results on this dataset separately in Table II together with a more suitable comparison group. Specifically, we compare the following methods: zero-shot CLIP\cite{radford2021learning}, which is evaluated without any task-specific data; few-shot CLIP-based methods, all evaluated with 1/2/4/8/16-shot setting, including CoOp\cite{zhou2022learning}, Tip-Adapter\cite{zhang2021tip}, SgVA-CLIP\cite{peng2023sgva}, CaFo\cite{zhang2023prompt}, GDA\cite{wang2024hard}, ProKeR\cite{bendou2025proker}, TransCLIP\cite{zanella2024boosting}, T-IDEA\cite{ye2025idea} and ProAPO\cite{qu2025proapo}; and our training-free UniFGVC, which is also evaluated under 1/2/4/8/16-shot setting but without any model tuning. Overall, UniFGVC achieves the best average accuracy of 84.2\%, surpassing the state-of-the-art CLIP-based method CaFo (78.66\%) by 4.62\%. In particular, UniFGVC achieves 81.1\% on ImageNet, surpassing CaFo by 12.3\%; 94.6\% on StanfordCars, with a margin of 18.3\%; and 61.1\% on FGVC-Aircraft, exceeding CaFo by 12.2\%. These consistent gains highlight the effectiveness of UniFGVC in capturing fine-grained semantics under limited data conditions. Notably, UniFGVC does not achieve the best performance on certain datasets such as EuroSAT and Caltech101, which we further analyze in Section~\ref{sec:non-object}. Overall, UniFGVC augments visual representations with structured semantic cues, thereby achieving more robust generalization and superior performance across diverse benchmarks.

Table~\ref{tab:2} presents the classification performance of recent fully-supervised MLLM-based FGVC methods on 5 datasets, including Flowers102\cite{nilsback2008automated}, OxfordPets\cite{parkhi2012cats}, FGVCAircraft\cite{maji2013fine}, StanfordCars\cite{krause20133d} and CUBbirds\cite{welinder2010caltech}. We compare with representative approaches such as LLaVA-Next\cite{liu2024improved}, FineR\cite{liu2024democratizing}, Idefics2\cite{laurenccon2024matters}, CascadeVLM\cite{wei2024enhancing} and Finedefics\cite{he2025analyzing}. Specifically, UniFGVC achieves an average accuracy of 83.21\%, surpassing the previous state-of-the-art method Finedefics (77.63\%) by 5.58\%. In particular, it reaches 95.74\% on Flowers and 77.48\% on CUBbirds, with absolute improvements of 5.86\% and 19.87\%. 

These results demonstrate that UniFGVC can achieve highly competitive performance even under substantially weaker task-specific supervision.
It is worth noting that this comparison involves different supervision regimes: the compared MLLM-based methods are fully supervised, whereas UniFGVC uses only few-shot support samples without parameter updates. A possible explanation for the competitive performance is that UniFGVC preserves the pretrained open-world knowledge of the underlying MLLM, avoiding the potential overfitting and knowledge degradation caused by adapting large models to limited task-specific data. Moreover, the CDV-Captioner explicitly elicits discriminative fine-grained attributes through reference-guided reasoning, allowing the few-shot samples to serve as task-specific retrieval anchors rather than optimization targets. Thus, Table II should be interpreted as demonstrating the competitiveness of the training-free retrieval paradigm under weaker supervision, rather than as a strictly controlled comparison of data efficiency.


\begin{table}[t]
    \centering
    \caption{Ablation study of CDV-Captioner on ImageNet under 1/2/4/8/16-shot settings. Different variants are evaluated by varying the description structure and reference guidance.
    }
    \label{tab:4}
    \begin{tabular}{lcccccc}\toprule
    Setting & 1 & 2 & 4 & 8 & 16 & avg. \\\midrule
    Image-Only & 37.88 & 45.76 & 52.18 & 57.14 & 60.72 & 50.74  \\
    Raw-Desc & 58.38 & 66.52 & 67.28 & 70.94 & 71.56 & 66.94 \\
    Struct-Desc & 65.42 & 70.96 & 73.66 & 74.20 & 76.20  & 72.09 \\
    Sum-Desc & 66.27 & 71.84 & 75.31 & 76.26 & 78.11  & 73.56 \\
    Random-Ref & 69.16 & 73.12 & 76.64 & 79.08 & 80.24 & 75.65 \\
    Similar-Ref & 70.18 & 74.42 & 78.00 & 80.24 & 81.08 & 76.78 \\\bottomrule
    \end{tabular}
\end{table}

\subsection{Effectiveness of CDV-Captioner}\label{sec:eff}

To thoroughly assess the contributions of different components within CDV-Captioner, we perform an ablation study on the ImageNet dataset under 1/2/4/8/16-shot settings, as shown in Table~\ref{tab:4}. We evaluate the following incremental configurations: \textbf{Image-Only} serves as the baseline, where retrieval relies solely on raw visual features without any textual guidance; \textbf{Raw-Desc} augments the baseline by incorporating naive textual descriptions generated by MLLM, which provide coarse semantic cues but lack structured attribute grounding; \textbf{Struct-Desc} further introduces region-based attribute descriptions, allowing the model to focus on discriminative local cues, though these are not guided by reference exemplars; \textbf{Sum-Desc} improves upon Struct-Desc by replacing multiple region-level descriptions with a concise summary of attributes, which enhances semantic compactness but still omits exemplar-based grounding; \textbf{Random-Ref} extends the framework with contrastive prompting using randomly selected reference exemplars, thereby injecting inter-class comparison signals, though in a noisy and less stable manner; \textbf{Similar-Ref} represents the full CDV-Captioner setup, where reference exemplars are chosen from visually similar categories, providing precise contrastive guidance and enabling structured comparative reasoning.

The ablation study reveals a clear progression of improvements as additional components are introduced. Naive textual descriptions already provide strong complementary cues over visual features, while structured and summarized region-based descriptions further refine semantic precision. Random references offer modest benefits, but visually similar references deliver consistent improvements by enabling precise comparative reasoning. These findings indicate that performance gains originate from two complementary sources: progressively refined textual descriptions, which enhance the richness and compactness of semantic cues, and the structured prompt guided by reference exemplars in CDV-Captioner, which injects category-aware contrastive signals. Together, these elements allow MLLM to extract highly discriminative fine-grained attributes, ultimately leading to robust recognition improvements under few-shot settings.

\begin{figure}
  \centering
  \includegraphics[width=\linewidth]{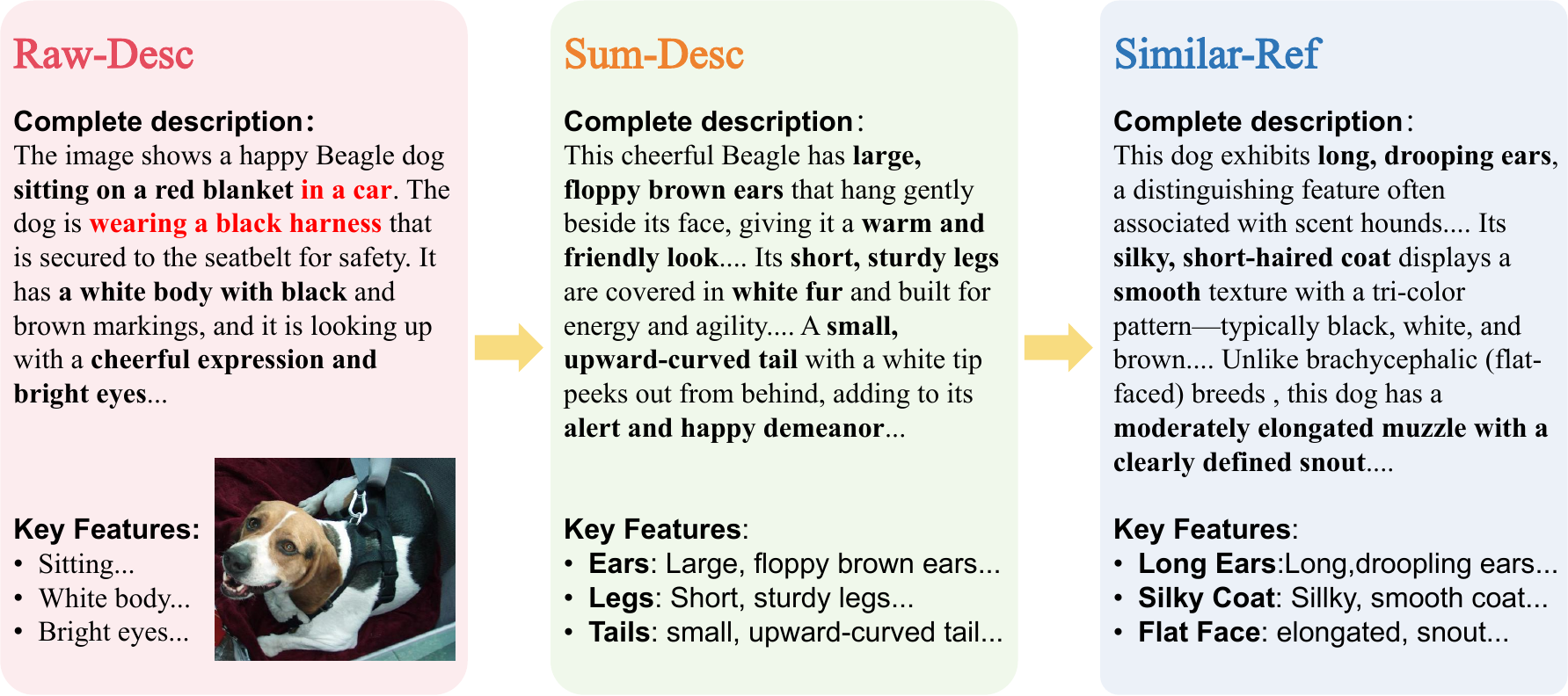}
  \caption{Visualization examples of structured attribute descriptions generated by Raw-Desc, Sum-Desc and Similar-Ref approaches.}
  \label{fig:ablation}
\end{figure}

\begin{figure}
  \centering
  \includegraphics[width=\linewidth]{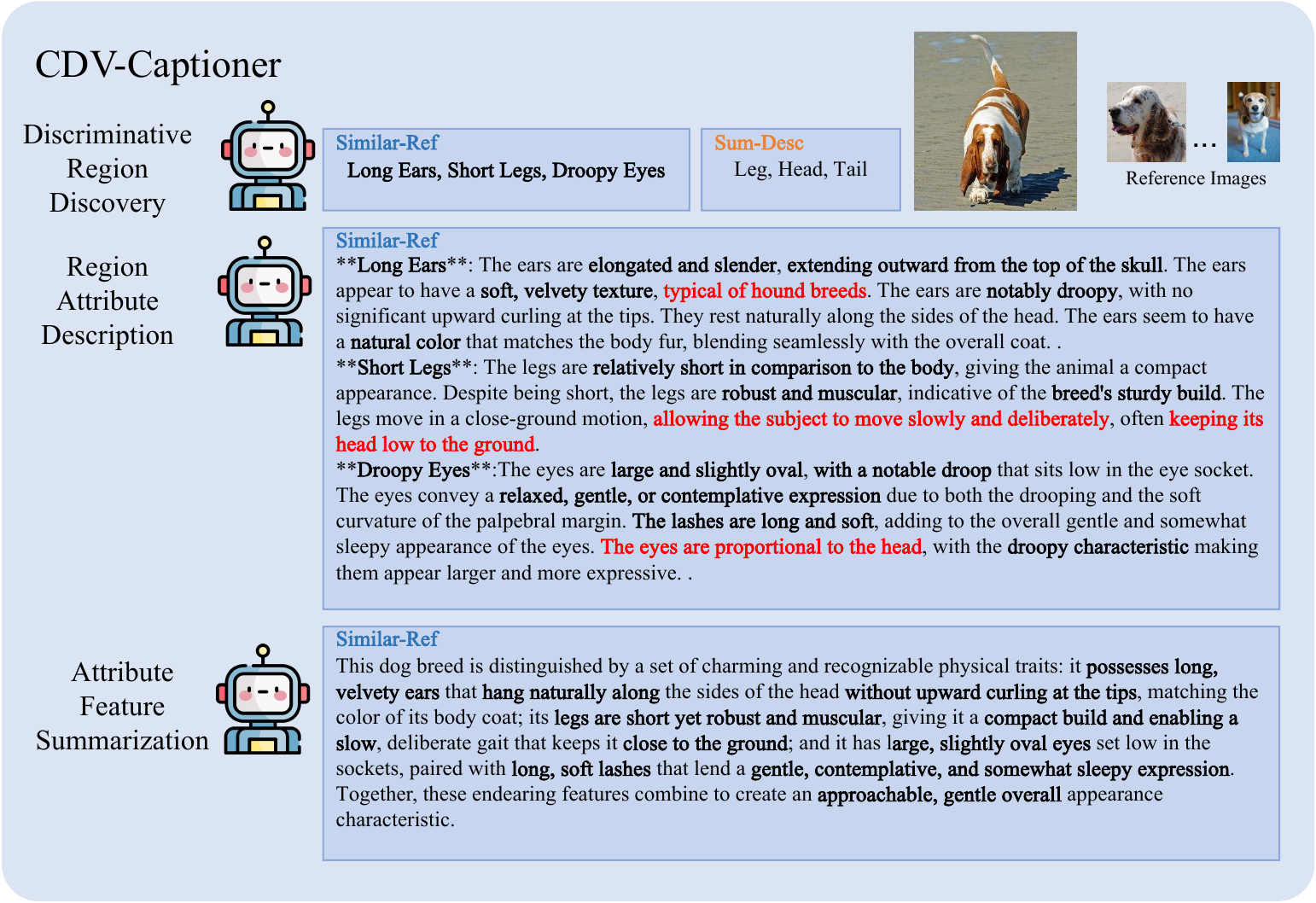}
  \caption{Visualization of the step-by-step refinement of the structured attribute description generated by CDV-Captioner.}
  \label{fig:COT-CDV-Captioner}
\end{figure}

Figure~\ref{fig:ablation} presents qualitative comparisons of structured descriptions generated under three configurations: \textbf{Raw-Desc}, \textbf{Sum-Desc}, and \textbf{Similar-Ref}. \textbf{Raw-Desc} tends to produce generic, scene-level descriptions with redundant and category-irrelevant details, which not only leads to coarse attributes lacking fine-grained discriminability but also increases the risk of hallucinatory content. \textbf{Sum-Desc} alleviates this issue by decomposing images into region-level attributes, enabling more focused part-based descriptions, yet the generated content remains relatively generic without reference guidance. In contrast, \textbf{Similar-Ref} combines region-level decomposition with exemplar-guided prompting, highlighting subtle but highly discriminative cues critical for fine-grained category differentiation while reducing hallucination. Figure~\ref{fig:COT-CDV-Captioner} further visualizes this refinement process inside CDV-Captioner: reference-guided reasoning first localizes more contrastive regions, then grounds finer-grained attribute descriptions on these regions, and finally summarizes them into a compact representation that preserves the core discriminative semantics while filtering redundant content. Together, these visualizations demonstrate a clear progression from coarse scene descriptions to structured, grounded, and reference-guided attribute representations, validating the effectiveness of CDV-Captioner in capturing fine-grained visual semantics, and a more detailed analysis of hallucination sensitivity is provided in Section~\ref{sec:hallucination}.

\subsection{Hyper-parameter Analysis}
In this section, we conduct ablation studies on UniFGVC with a focus on analyzing the impact of key hyper-parameters on performance, using the ImageNet dataset as a case study.

\textbf{Number of regions.} To assess the effect of structured discriminative regions, we vary the number of regions
\begin{math}
    \ s \in \left \{1,2,3,4,5\right \} 
\end{math}, as shown in Table~\ref{tab:3}. Increasing \textit{s} from 1 to 3 consistently improves retrieval performance by enabling richer visual–semantic representations, while further increasing \textit{s} leads to diminishing returns, indicating performance saturation once sufficient regions are captured. Larger values of \textit{s} also lead to longer descriptions, increasing the risk of irrelevant or hallucinatory details and reducing efficiency. Balancing accuracy and efficiency, we set
\begin{math}
    \ s = 3
\end{math} in all subsequent experiments.


\begin{table}[t]
    \centering
    \caption{Ablation study on the number of regions (s) under the few-shot setting on ImageNet.}
    \label{tab:3}
    \begin{tabular}{ccccccc}\toprule
        s  & 1 & 2 & 4 & 8 & 16 & avg.\\\midrule
        1 & 66.24 & 67.22 & 71.15 & 72.28 & 73.62 & 70.10 \\
        2 & 67.25 & 71.08 & 74.97 & 77.44 & 78.33 & 73.81 \\
        3 & 70.18 & 74.42 & 78.00 & 80.24 & 81.08 & 76.78 \\
        4 & 70.24 & 74.44 & 77.56 & 80.83 & 80.62 & 76.74 \\
        5 & 70.21 & 74.32 & 77.98 & 81.00 & 82.18 & 77.14 \\\bottomrule
    \end{tabular}
\end{table}

\begin{table}[t]
    \centering
    \caption{Ablation study on the number of reference samples (t) under the few-shot setting on ImageNet.}
    \label{tab:reference}
    \begin{tabular}{ccccccc}\toprule
        t & 1 & 2 & 4 & 8 & 16 & avg. \\\midrule
        0 & 66.27 & 71.84 & 75.31 & 76.26 & 78.11 & 73.56 \\
        1 & 69.03 & 74.30 & 77.76 & 79.38 & 81.33 & 76.36 \\
        2 & 69.25 & 74.42 & 77.04 & 80.23 & 80.32 & 76.25 \\
        3 & 70.63 & 74.68 & 77.34 & 80.21 & 80.20 & 76.61 \\
        4 & 70.18 & 74.42 & 78.00 & 80.24 & 81.08 & 76.78 \\\bottomrule
    \end{tabular}
\end{table}

\textbf{Number of reference samples.} To assess the influence of the number of reference samples on fine-grained recognition performance, we conduct an ablation study with \begin{math}
    \ t \in \left \{0,1,2,3,4\right \} 
\end{math}. As shown in Table~\ref{tab:reference}, performance consistently improves as \textit{t} increases,demonstrating the effectiveness of incorporating reference samples for contrastive reasoning. The setting with \begin{math}
    \ t = 0 
\end{math} corresponds to the \textbf{Sum-Desc}, while for \begin{math}
    \ t > 0
\end{math}, the CDV-Captioner integrates reference samples to enable contrastive generation, leading to consistent improvements across few-shot settings. Notably, performance is not highly sensitive to the exact value of \textit{t}: even a single exemplar yields substantial gains, and using four reference samples provides more stable improvements.

\begin{table}[t]
    \centering
    \caption{ablation study on the compatibility of UniFGVC with different vision encoders under the few-shot setting on ImageNet.}
    \label{tab:5}
    \begin{tabular}{ccccccc}\toprule
        Encoders & 1 & 2 & 4 & 8 & 16 & avg. \\\midrule
        Bge-m3 & 57.90 & 59.02 & 62.28 & 66.90 & 67.44 & 62.71 \\
        CLIP  & 59.08 & 62.92 & 66.08 & 66.94 & 67.82 & 64.57 \\
        RADIO  & 64.64 & 68.96 & 70.38 & 71.56 & 72.60 & 69.63 \\
        Unicom  & 70.18 & 74.42 & 78.00 & 80.24 & 81.08 & 76.78 \\\bottomrule
    \end{tabular}
\end{table}

\begin{table}[t]
    \centering
    \caption{ablation study on the compatibility of UniFGVC with different MLLM under the few-shot setting on ImageNet.}
    \label{tab:6}
    \begin{tabular}{ccccccc}\toprule
        MLLM & 1 & 2 & 4 & 8 & 16 & avg. \\\midrule
        InternVL-8B & 60.17 & 66.10 & 70.94 & 72.62 & 74.70 & 68.91 \\
        QwenVL-2B & 66.46 & 70.31 & 74.23 & 75.99 & 78.07 & 73.01 \\
        QwenVL-7B  & 70.18 & 74.42 & 78.00 & 80.24 & 81.08 & 76.78 \\
        GPT-4o & 73.21 & 76.54 & 81.06 & 82.31 & 84.03 & 79.43\\\bottomrule
    \end{tabular}
\end{table}

\begin{table}[t]
    \centering
    \caption{Comparison of different multimodal fusion strategies on ImageNet.}
    \label{tab:8}
    \begin{tabular}{ccccccc}\toprule
        Setting & 1 & 2 & 4 & 8 & 16 & avg. \\\midrule
        Max pooling & 64.23 & 65.88 & 70.91 & 72.34 & 73.12 & 69.15 \\
        Averaging & 64.36 & 65.68 & 71.06 & 72.31 & 74.03 & 69.49\\
        Concatenation & 66.46 & 69.72 & 74.69 & 76.21 & 77.48 & 72.91\\
        Hybridization & 67.48 & 70.92 & 75.31 & 76.20 & 77.62 & 73.51 \\\bottomrule
    \end{tabular}
\end{table}

\begin{table}[t]
    \centering
    \caption{Hallucination scores of CDV-Captioner variants on ImageNet.}
    \label{tab:hall}
    \begin{tabular}{cccc}\toprule
        Setting & MLLM & Hall. Score & Hall. Level\\\midrule
        Raw-Desc &  \multirow{3}{*}{\centering Qwen2.5-VL-7B} & 67.37 & Medium \\
        Sum-Desc & & 81.72 & Low \\
        Similar-Ref & & 82.04 & Low \\
        \midrule
        Raw-Desc & \multirow{3}{*}{\centering GPT-4o} & 78.62 & Low \\
        Sum-Desc & & 81.94 & Low \\
        Similar-Ref & & 82.07 & Low \\
        \bottomrule
    \end{tabular}
\end{table}

\subsection{Compatibility of UniFGVC}
\textbf{Different encoders.} In Table~\ref{tab:5}, we assess the compatibility of UniFGVC with different encoders. For fair comparison, all visual encoders share a similar architectural scale (ViT-L/14), while textual encoding is fixed using Jina-CLIP-V2\cite{koukounas2024jina}. We evaluate four distinct visual encoders: Unicom\cite{an2023unicom}, RADIO\cite{ranzinger2024radio}, CLIP\cite{radford2021learning} and Bge-m3\cite{chen2024bge}. Notably, Bge-m3 employs aligned vision and text encoders, while the others use independent components. The results highlight the strong adaptability of UniFGVC across diverse encoder backbones. Even with the standard CLIP\cite{radford2021learning}, our framework achieves competitive performance, while pairing with stronger encoders such as Unicom\cite{an2023unicom} further boosts accuracy. Notably, the aligned Bge-m3\cite{chen2024bge} variant performs less effectively compared to Unicom\cite{an2023unicom}, suggesting that UniFGVC benefits more from the complementary strengths of independently optimized encoders rather than strict modality alignment. Overall, these findings confirm that UniFGVC can seamlessly integrate with different encoder architectures and consistently deliver robust performance, underscoring its versatility as a training-free universal FGVC solution.



\textbf{Different MLLM.} To evaluate the generalizability and plug-and-play flexibility of UniFGVC, we conduct an ablation study with four different MLLM, including  InternVL-8B\cite{chen2024far}, Qwen2.5-VL-2B\cite{bai2023qwen}, Qwen2.5-VL-7B, and GPT-4o\cite{achiam2023gpt}. As shown in Table~\ref{tab:6}, UniFGVC consistently achieves competitive performance across all models, indicating that it is not tied to any specific architecture. Notably, we observe clear performance variance across different MLLM. This can be attributed to three factors: (1) Model scale and visual encoder quality—larger models with advanced vision towers better localize discriminative regions; (2) Pre-training data for fine-grained tasks—closed-source models like GPT-4o benefit from richer fine-grained annotations; (3) Chain-of-Thought following ability—stronger MLLM execute multi-step prompts (region discovery → attribute description → summarization) more faithfully. Despite the variance, UniFGVC treats the MLLM as a pluggable module only for description generation. The core retrieval pipeline (encoders, gallery, similarity matching) is MLLM-agnostic. Users can freely substitute the MLLM based on deployment constraints: lightweight models (e.g., Qwen2.5-VL-2B) for edge devices, or powerful models (e.g., GPT-4o) for maximum accuracy. This plug-and-play design is a key advantage of our training-free framework. What’s more, as evidenced by Table~\ref{tab:hall}, our CDV-Captioner can effectively shrinks the performance gap between MLLM in attribute descriptions generation, demonstrating that our structured, reference-guided prompting successfully mitigates the dependency on the underlying MLLM’s intrinsic capabilities.

\textbf{Different fusion strategies.} 
To further examine the role of multimodal feature integration, we analyze several fusion strategies for combining visual and textual features. Specifically, we consider: 1) \textbf{Averaging}, which merges the two modalities by mean pooling to obtain a stable but coarse representation; 2) \textbf{Max pooling}, which selects the element-wise maximum values across the two modalities; 3) \textbf{Concatenation}, which directly concatenates visual and textual features to preserve modality-specific information in a unified embedding space; 4) \textbf{Hybridization}, inspired by Bge-m3\cite{chen2024bge}, which combines complementary retrieval views by re-ranking candidates with weighted scores from multiple feature spaces, but requires additional training. As shown in Table~\ref{tab:8}, we evaluate these strategies on ImageNet using the visual encoder and text encoder from Bge-m3. The results demonstrate that although \textbf{Hybridization} achieves the best performance, it often relies on specially designed or trained fusion mechanisms. In contrast, \textbf{Concatenation} requires no additional training yet delivers consistently strong results. Furthermore, its performance improves with the capability of the underlying encoders.


\begin{table*}[t]
    \centering
    \caption{Efficiency analysis for different methods on ImageNet. CLIP-based methods are evaluated under the 16-shot setting, MLLM-based methods under the fully-supervised setting, and UniFGVC under the 16-shot setting. For UniFGVC, the reported training time corresponds to the time required for multimodal category template construction.}
    \label{tab:effic}
    \begin{tabular}{lccccc}\toprule
        Models & Setting & Accuracy & Training Time & Infer. speed/Image & Trainable Params \\\midrule
        CoOp & \multirow{3}{*}{\centering 16-shot} & 62.95 & 205min & 6.90ms & 2028 \\
        Tip-Adapter & & 65.44 & 198min & 6.52ms & 16M \\
        SgVA-CLIP & & 65.80 & 218min & 7.05ms & 16M \\\midrule
        Finedefics & \multirow{2}{*}{\centering fully-supervised} & 75.32 & 672min & 14.45ms & 18M \\
        CascadeVLM &  & 80.61 & 646min & 18.36ms & 32M \\
        \midrule
        \textbf{UniFGVC} & 16-shot & 81.08 & 186min & 18.11ms & 0 \\\bottomrule
    \end{tabular}
\end{table*}

\begin{table}[t]
    \centering
    \caption{Robustness evaluation under injected hallucination on ImageNet.}
    \label{tab:robu}
    \begin{tabular}{ccc}\toprule
        Hall. Injection & Hall. Level & Accuracy\\\midrule
        0 & Low & 81.08 \\
        1 & Low & 80.91 \\
        2-3 & Medium & 80.27 \\
        4-8 & High & 79.87 \\\bottomrule
    \end{tabular}
\end{table}

\subsection{Robustness to Hallucination}\label{sec:hallucination}
Although structured textual descriptions enhance feature discriminability, MLLM-generated content may introduce hallucinations that are irrelevant or inconsistent with the target class. To evaluate the robustness of UniFGVC against such hallucinations, we conduct two complementary analyses.

\textbf{Hallucination Probability of CDV-Captioner.} We assess the hallucination tendency of different description strategies, including \textbf{Raw-Desc}, \textbf{Sum-Desc} and \textbf{Similar-Ref}. Following recent evaluation protocols, we adopt a hybrid evaluation strategy using GPT-4o\cite{achiam2023gpt} as the automatic evaluator and human verification for validation. The scoring criteria for the generated descriptions are detailed as follows:
\begin{itemize}
\item \begin{math}
  \  > 90:
\end{math} Detailed category-relevant description without additional irrelevant information.
\item \begin{math}
  \  80-89:
\end{math} Minor off-category information but no hallucination.
\item \begin{math}
  \  70-79:
\end{math} Minor off-category information with limited hallucination.
\item \begin{math}
  \  60-69:
\end{math} Considerable off-category information with hallucination.
\item \begin{math}
  \  < 60:
\end{math} Severe hallucination with large amounts of irrelevant information.
\end{itemize}

The final hallucination score for each setting is the average of scores over all test samples. To validate the automatic scoring, we randomly sampled 20\% of the descriptions (balanced across categories) and had two human annotators rate them using the same criteria. The Cohen’s Kappa between GPT-4o and the average human rating was 0.62, which is comparable to human annotators (inter-human Kappa on the same sample is 0.65) and indicates substantial agreement according to \cite{landis1977measurement}. This confirms the reliability of our automatic evaluation.

We further evaluate the description generation quality of Qwen2.5-VL-7B, a representative lightweight open-source MLLM, and GPT-4o, a strong closed-source MLLM, under three settings (Raw-Desc, Sum-Desc, and Similar-Ref) on the full ImageNet test set. We choose Qwen2.5-VL-7B as the main backbone because it provides a practical balance between reproducibility, deployment efficiency, and multimodal reasoning capability.

As shown in Table~\ref{tab:hall}, under the basic Raw-Desc setting, Qwen2.5-VL-7B exhibits a clear gap in hallucination mitigation compared with GPT-4o. However, after applying our Sum-Desc and Similar-Ref mechanisms, the hallucination score of Qwen2.5-VL-7B is significantly improved, bringing its hallucination-control performance close to that of GPT-4o under the evaluated settings. This result indicates that CDV-Captioner substantially narrows the gap in description faithfulness between a lightweight open-source MLLM and a strong closed-source MLLM. This improvement is primarily attributed to two factors. First, reference-guided chain-of-thought reasoning helps the model localize category-discriminative regions and generate attribute descriptions grounded in visible evidence, which suppresses class-irrelevant or fabricated content. Second, the summarization step re-accesses the original image to verify and refine region-level outputs, filtering inconsistent or noisy descriptions. Together, these mechanisms produce more faithful, reliable, and robust textual representations.

\begin{table}[t]
    \centering
    \caption{Retrieval paradigm comparison on ImageNet, where “Retr” denotes per-image retrieval latency and “Infer” denotes end-to-end inference latency (ms).}
    \label{tab:retrieval}
    \begin{tabular}{lccccc}\toprule
        Retrieval Paradigm & mAP & R@10 & Accuracy & Retr & Infer \\\midrule
        Nearest-Neighbor & 64.28 & 76.87 & 81.08 & 8.21 & 19.38 \\
        Clustering-Based & 60.16 & 75.44 & 77.03 & 6.03 & 12.71 \\
        Hash-Based & 56.33 & 70.57 & 71.80 & 4.83 & 8.43  \\\bottomrule
    \end{tabular}
\end{table}

\textbf{Performance robustness under injected hallucination.} To further assess the robustness of our framework against hallucination, we conduct an additional experiment on ImageNet under 16-shot setting by manually injecting hallucinated attributes into the textual descriptions while keeping visual features unchanged. As shown in Table~\ref{tab:robu}, we progressively increase the degree of injected hallucination from 1 attribute, to 2-3 attributes, and up to 4-8 attributes. The results show only a slight accuracy drop, indicating that the framework remains highly stable even under strong noise. This robustness primarily stems from our retrieval-based design, where the fused multimodal representations are anchored by visual features, effectively mitigating the negative influence of hallucinated or inaccurate textual content.

\begin{figure}
  \centering
  \includegraphics[width=\linewidth]{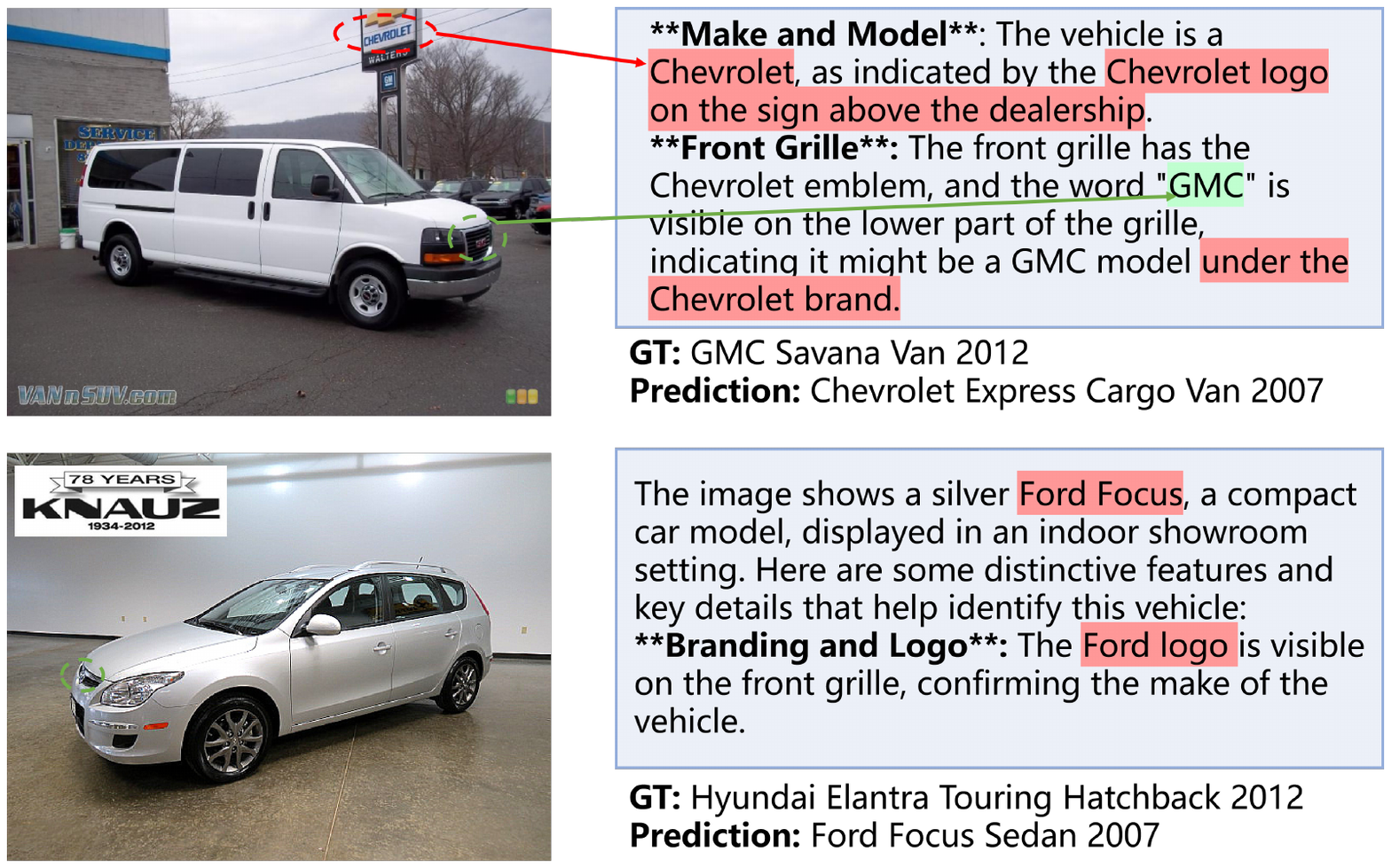}
  \caption{ Qualitative failure cases of UniFGVC caused by MLLM hallucination or attribute misinterpretation. Erroneous textual cues are highlighted in red, while the corresponding correct visual evidence is highlighted in green.}
  \label{fig: hallucination failure case}
\end{figure}

\textbf{Qualitative failure analysis}. To complement the quantitative hallucination evaluation, Fig. \ref{fig: hallucination failure case} presents two representative failure cases where incorrect MLLM-generated descriptions lead to recognition errors. These cases show that severe hallucinations involving category-specific visual cues can still propagate to the final prediction, despite the overall robustness of the retrieval-based framework. This analysis highlights an important limitation of UniFGVC: the proposed CDV-Captioner reduces hallucination on average but does not completely eliminate errors caused by severe visual misattribution or incorrect fine-grained attribute interpretation.


\begin{figure}
  \centering
  \includegraphics[width=\linewidth]{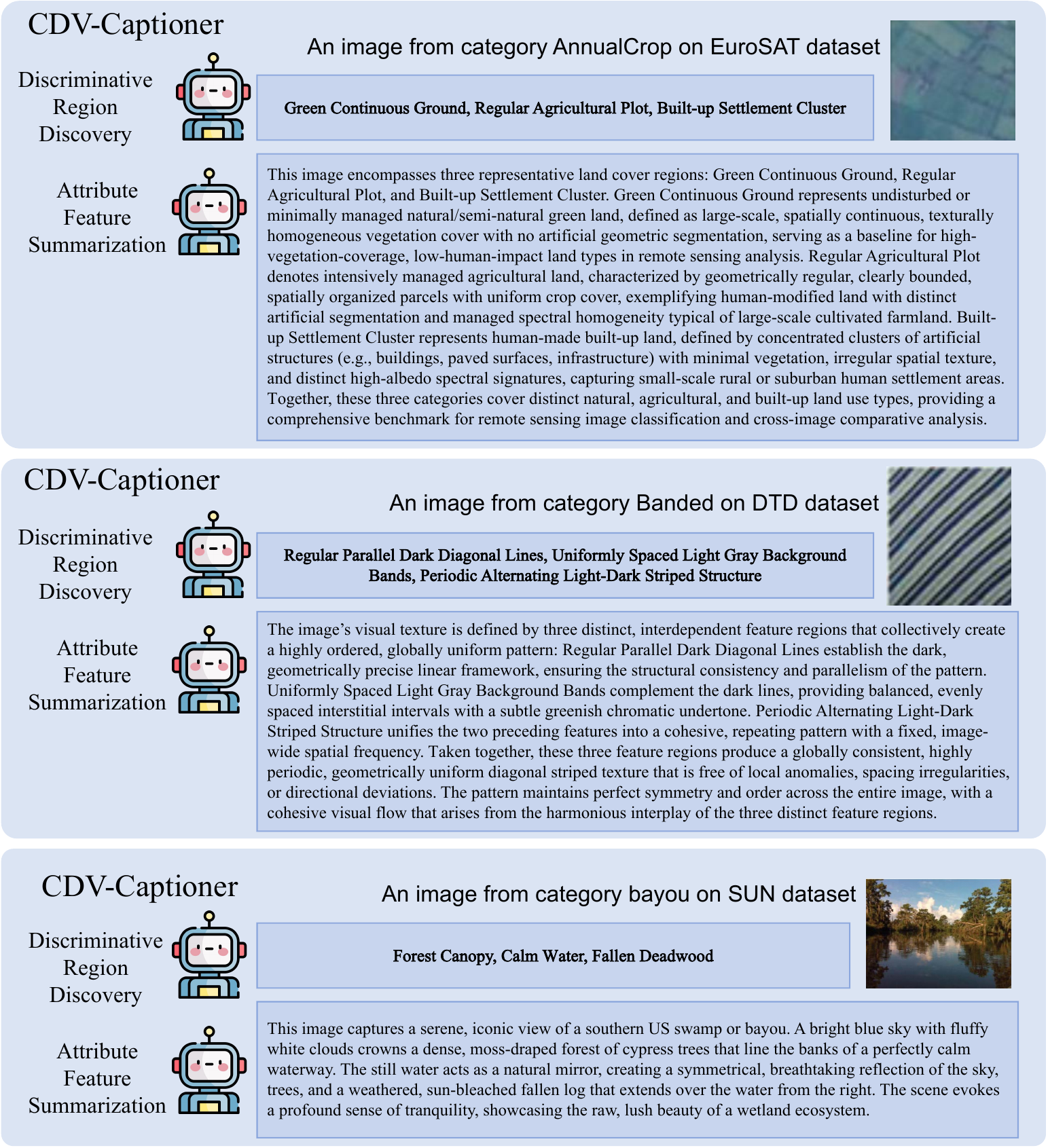}
  \caption{Visualization of CDV-Captioner on scene-centric and non-object datasets including EuroSAT, DTD, and SUN.}
  \label{fig:euro}
\end{figure}

\subsection{Efficiency and Retrieval Scalability Analysis}
To further evaluate the efficiency of UniFGVC, we compare its computational cost with representative CLIP-based and MLLM-based methods in Table X. For a fair comparison, we report both the offline template construction cost and the online inference latency. Specifically, UniFGVC requires 186 min to construct the multimodal category templates offline. This is a one-time preprocessing cost: the generated descriptions and corresponding image/text features are cached and reused for subsequent queries. Moreover, when new categories are introduced, only the templates of the newly added categories need to be constructed and appended to the existing gallery, without reprocessing previously constructed categories.

During online inference, UniFGVC takes 18.11 ms/query in total, including 13.64 ms for CDV-Captioner generation and 4.471 ms for feature extraction, processing, and retrieval. The latter consists of 3.369 ms for text feature extraction, 0.258 ms for image feature extraction, 0.440 ms for processing, normalization, and fusion, and only 0.404 ms for retrieval. These results indicate that the MLLM-based caption generation is the dominant source of online computational cost, while multimodal feature processing and retrieval introduce relatively small overheads.

Despite the additional online latency compared with conventional CLIP-based methods, UniFGVC avoids task-specific model training and supports incremental extension to novel categories with only category-level template construction. Furthermore, the online cost can be further reduced by adopting a smaller MLLM for caption generation or a coarse-to-fine retrieval strategy, where visual features are first used for efficient candidate retrieval and the more expensive CDV-Captioner-based reasoning is applied only to a small top-\(k\) candidate set. These properties make UniFGVC particularly suitable for few-shot scenarios where flexibility and adaptation to changing category sets are important.

To analyze the scalability of UniFGVC in large-scale applications, we further study different retrieval paradigms on ImageNet-1K, including exact nearest-neighbor retrieval, clustering-based retrieval, and hash-based retrieval, with results reported in Table~\ref{tab:retrieval}. In the exact setting, UniFGVC retrieves over a category template library with complexity O(CK), where C is the number of categories and K is the number of templates per category, which may become increasingly expensive as the category space grows. To provide a more complete evaluation, we report not only final classification accuracy, but also retrieval-oriented metrics including mean Average Precision (mAP) and Recall@10 (R@10), together with search time and end-to-end inference time.

The results show that exact nearest-neighbor retrieval achieves the best retrieval quality and classification performance, but incurs the highest search cost. Clustering-based retrieval reduces the search space and improves efficiency, at the cost of a moderate drop in retrieval quality. Hash-based retrieval further accelerates search through compact binary representations while still maintaining competitive performance. We also observe that the end-to-end speedup (refer to the Infer in Table~\ref{tab:retrieval}) is smaller than the retrieval-stage speedup (refer to the Retr in Table~\ref{tab:retrieval}), because the MLLM-based reasoning process for generating attribute-aware descriptions constitutes a substantial portion of the overall latency. More importantly, since UniFGVC is ultimately formulated as retrieval over a multimodal template library, it remains naturally compatible with mature large-scale retrieval infrastructures, including approximate indexing, compressed search, and hash-based retrieval systems.

\begin{table}[t]
    \centering
    \caption{Ablation study of CDV-Captioner on EuroSAT under 1/2/4/8/16-shot settings.}
    \label{tab:euro}
    \begin{tabular}{lcccccc}\toprule
    Setting & 1 & 2 & 4 & 8 & 16 & avg. \\\midrule
    Image-Only & 42.51 & 58.67 & 66.22 & 68.91 & 76.33 & 62.53  \\
    Raw-Desc & 43.21 & 60.05 & 67.26 & 72.52 & 76.69 & 63.95 \\
    Similar-Ref & 56.74 & 67.24 & 75.65 & 81.69 & 84.57 & 73.18 \\\bottomrule
    \end{tabular}
\end{table}

\subsection{Applicability to Non-Object and Scene-Centric Fine-Grained Recognition}\label{sec:non-object}

While Table I shows that UniFGVC achieves strong overall performance, its gains on scene-centric or non-object datasets such as EuroSAT, SUN, and DTD are less pronounced than on object-centric benchmarks. This difference suggests that the effectiveness of UniFGVC is influenced by how class-discriminative evidence is structured in the data.

In object-centric datasets, discriminative cues are often localized and naturally expressed as semantic parts or attributes. By contrast, scene-centric categories usually rely on more distributed evidence, such as texture regularities, spatial layouts, or environmental structures. Accordingly, the “regions” identified by CDV-Captioner should not be understood only as physical object parts, but more generally as localized discriminative visual evidence. As illustrated in Figure~\ref{fig:euro}, CDV-Captioner can still capture meaningful patterns in non-object scenarios, such as “green continuous ground” in EuroSAT, or “regular parallel dark lines” in DTD, showing that the region discovery mechanism remains applicable beyond object-centric recognition.

We further validate this observation on EuroSAT in Table~\ref{tab:euro}. Compared with the Image-Only baseline, both raw descriptions and reference-guided descriptions consistently improve performance across all shot settings, indicating that the proposed region discovery and attribute-aware description mechanism remains beneficial rather than harmful in non-object settings. Nevertheless, because scene-centric evidence is more distributed and less naturally verbalized as compact attribute-like descriptions, the current region-centric prompting strategy is more naturally aligned with object-centric fine-grained recognition. This also suggests a promising future direction toward layout-aware or texture-aware prompting strategies.

\section{Conclusion}
In this paper, we present UniFGVC, a training-free framework that reformulates fine-grained visual classification as a multimodal retrieval problem. By leveraging MLLM and reference-guided captioning, UniFGVC constructs discriminative image–text pairs from few-shot data. Extensive experiments demonstrate that UniFGVC outperforms CLIP-based few-shot methods and several fully supervised MLLM-based baselines. In future work, we plan to extend UniFGVC to more challenging fine-grained domains, exploring both broader category coverage and finer granularity to further advance training-free fine-grained recognition.

\section{Acknowledgements}
This work was supported by National Natural Science Foundation of China under Grant No.62276260.

\bibliographystyle{IEEEtran}
\bibliography{tmm}

\begin{IEEEbiography}[{\includegraphics[width=1in,height=1.25in,clip,keepaspectratio]{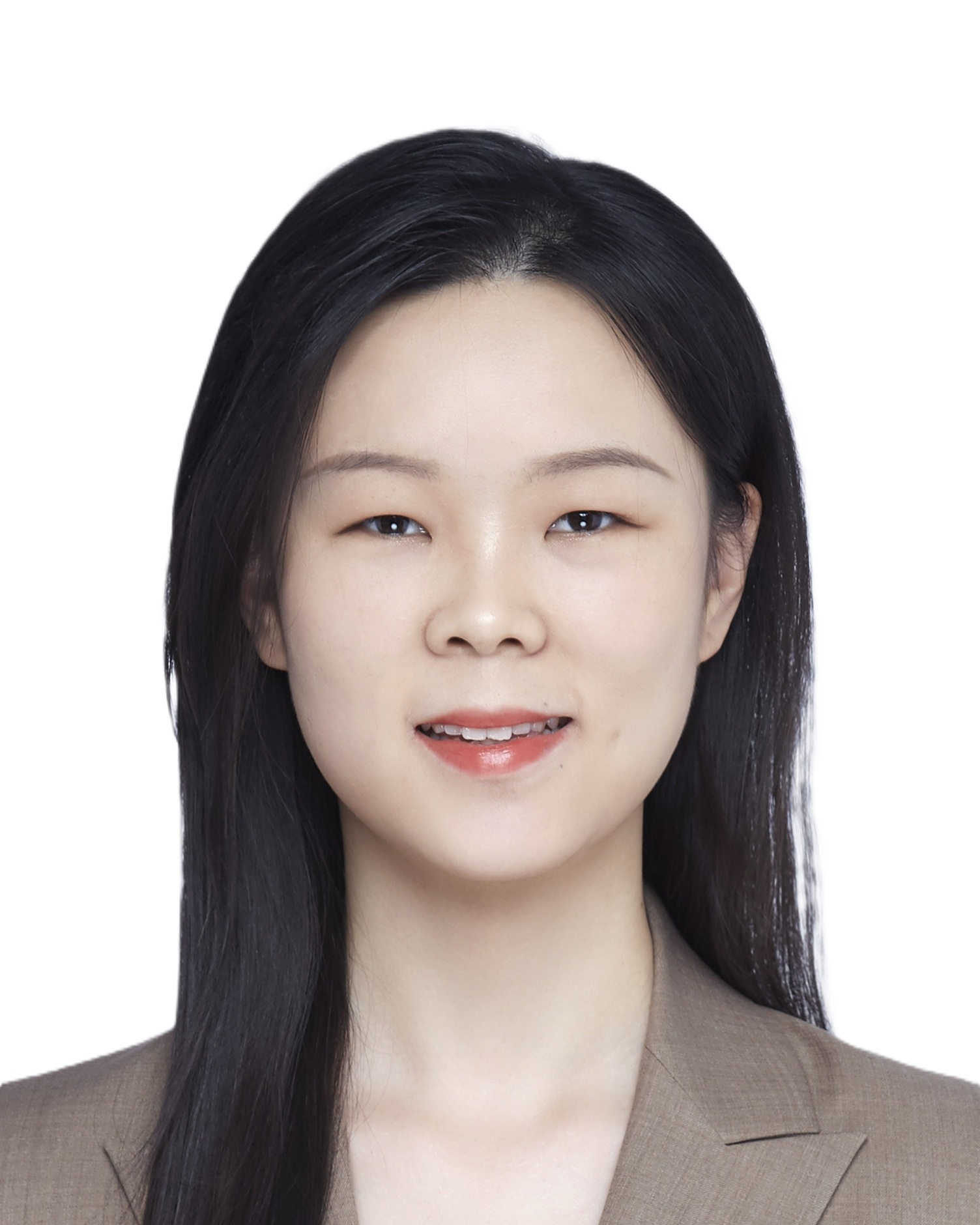}}]{Haiyun Guo}
received the B.S. degree from Wuhan University, Wuhan, China, in 2013, and the Ph.D. degree from the University of Chinese Academy of Sciences, Beijing, China, in 2018. She is currently an Associate Researcher with the Institute of Automation, Chinese Academy of Sciences, Beijing, China. She has authored or coauthored more than 40 papers in journals and conferences, including IEEE TIP, IEEE TMM, IEEE TNNLS, CVPR, ICML, NeurIPS, ECCV, EMNLP, ACL, and ACM Multimedia. Her research interests include fine-grained understanding and generation of multimodal content, continual learning of large models, and related areas in computer vision and machine learning.
\end{IEEEbiography}

\begin{IEEEbiography}[{\includegraphics[width=1in,height=1.25in,clip,keepaspectratio]{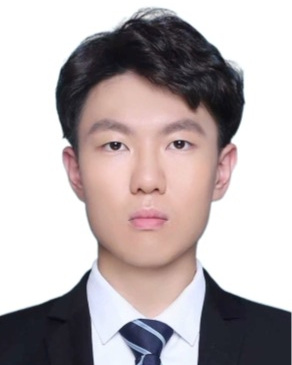}}]{Hongyu Guo}
received the B.S. degree from Dalian Maritime University, Dalian, China, in 2023. He is currently working toward the M.S. degree with Beijing Jiaotong University, Beijing, China. His research focuses on multimodal retrieval and multimodal large models, with a particular emphasis on fine-grained visual classification.
\end{IEEEbiography}

\begin{IEEEbiography}[{\includegraphics[width=1in,height=1.25in,clip,keepaspectratio]{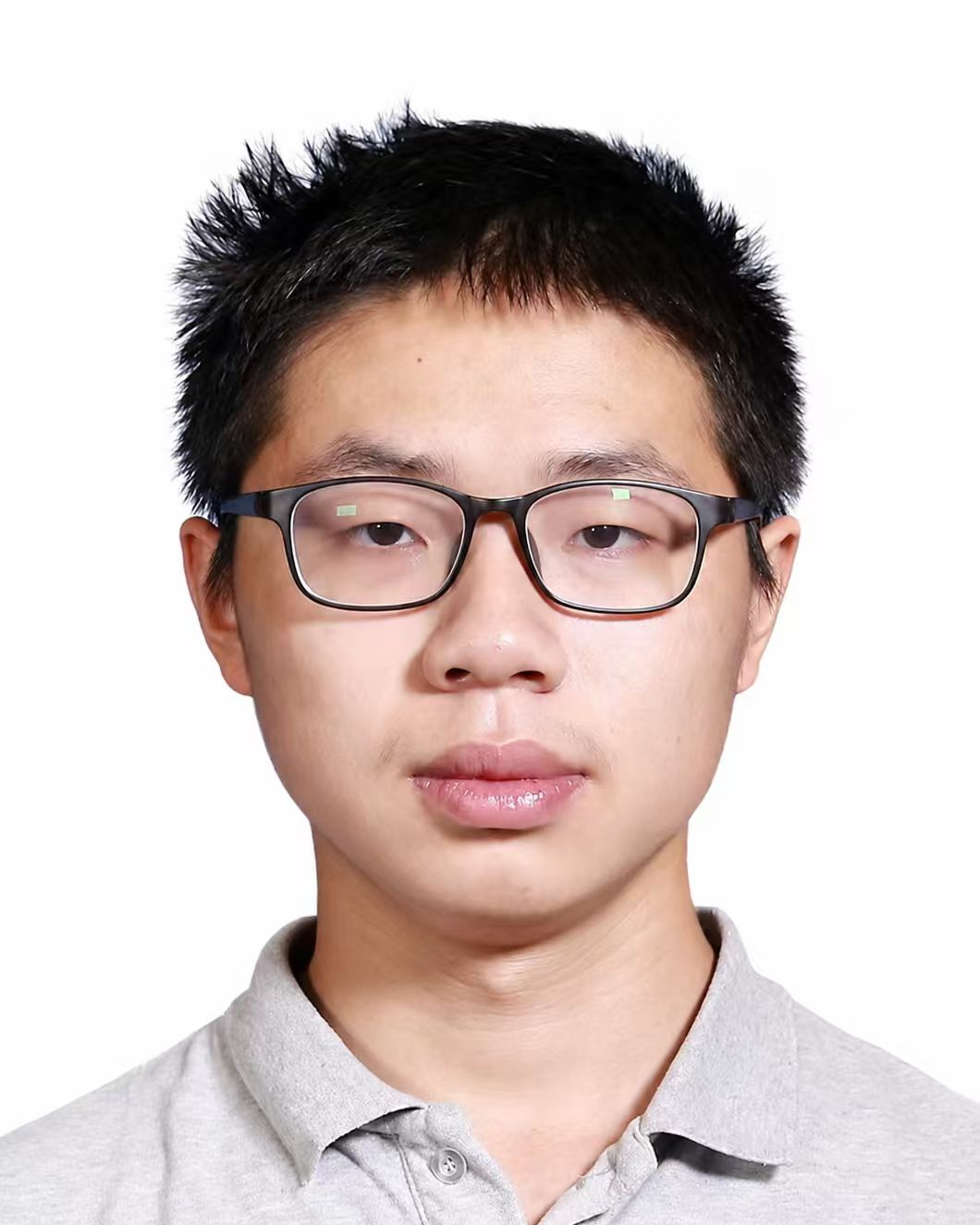}}]{Xiangzhao Hao}
received the B.S. degree from Tianjin University, Tianjin, China, in 2023, and the M.S. degree from the Institute of Automation, Chinese Academy of Sciences, Beijing, China, in 2026. He is currently pursuing the Ph.D. degree with the School of Artificial Intelligence, Shanghai Jiao Tong University, Shanghai, China, in a joint program with the Shanghai Innovation Institute. His research interests include unified multimodal models and universal multimodal embedding and retrieval.
\end{IEEEbiography}

\begin{IEEEbiography}[{\includegraphics[width=1in,height=1.25in,clip,keepaspectratio]{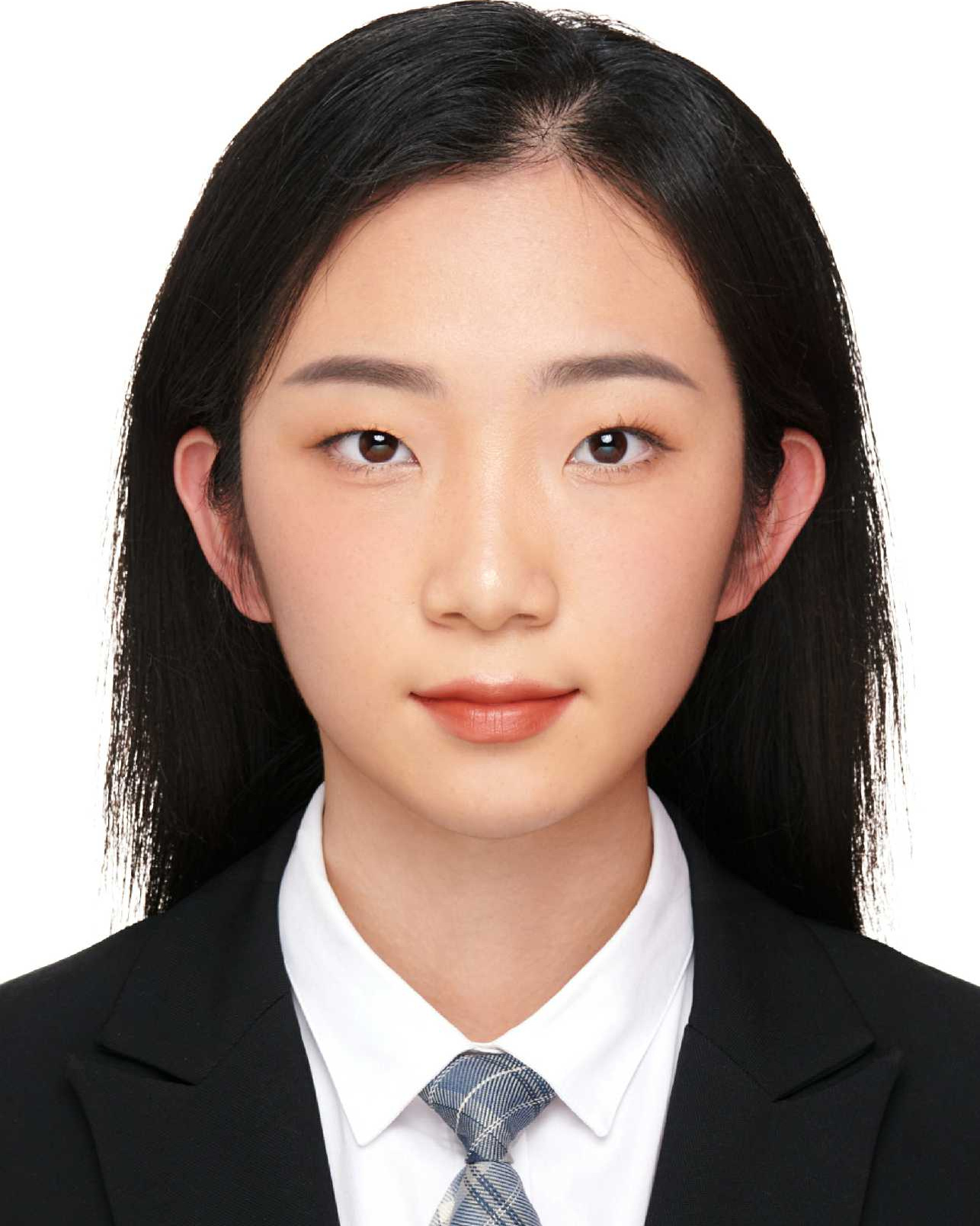}}]
{Jiarui Guo}
is currently a master's student in Computer Science at The University of Hong Kong. Her research interests include multimodal large language models, computer vision, and machine learning, particularly multimodal reasoning, image retrieval, fine-grained visual classification, and AI model robustness.
\end{IEEEbiography}

\begin{IEEEbiography}[{\includegraphics[width=1in,height=1.25in,clip,keepaspectratio]{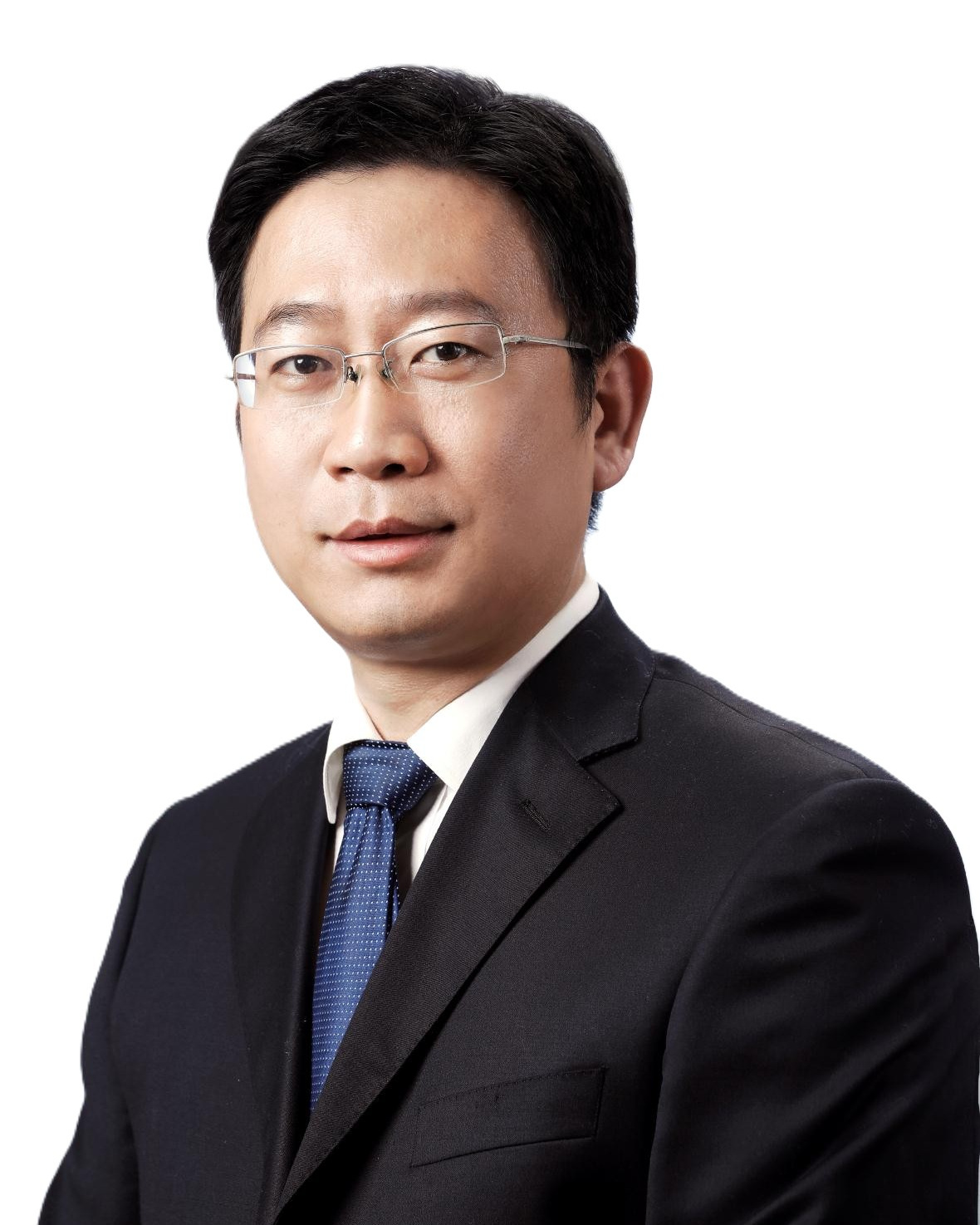}}]{Jinqiao Wang}
received the M.S. degree from Tianjin University, Tianjin, China, in 2004, and the Ph.D. degree from the Institute of Automation, Chinese Academy of Sciences, Beijing, China, in 2008. He is currently a Professor with the Institute of Automation, Chinese Academy of Sciences, Beijing, China. He has authored or coauthored more than 300 papers in leading journals and conferences, including IEEE TIP, IEEE TMM, IEEE TCSVT, CVPR, ICCV, and ACM Multimedia. His research interests include multimodal foundation models, fine-grained visual understanding, efficient learning and deployment of large models, and embodied artificial intelligence.
\end{IEEEbiography}

\begin{IEEEbiography}[{\includegraphics[width=1in,height=1.25in,clip,keepaspectratio]{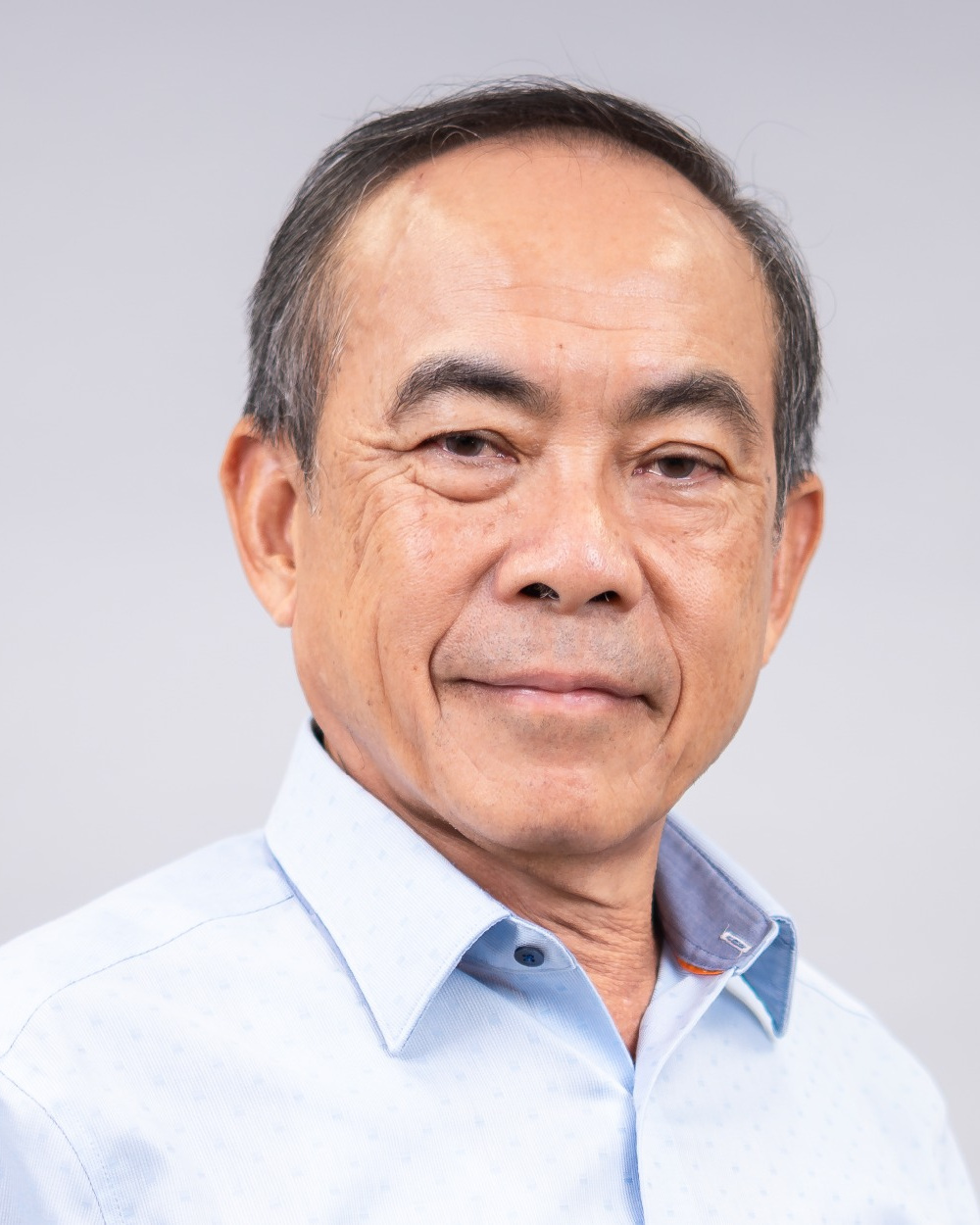}}]
{Tat-Seng Chua}
received the Ph.D. degree in computer science from the University of Leeds, Leeds, U.K., in 1983. He is currently the KITHCT Chair Professor with the School of Computing, National University of Singapore, Singapore, and a Distinguished Visiting Professor with Tsinghua University, Beijing, China. He is also the Co-Director of the NUS-Tsinghua Extreme Search Center (NExT). His research interests include multimodal foundation models, responsible artificial intelligence, and conversational search and recommendation. He received the ACM SIGMM Achievements Award in 2015 and the CCF Overseas Outstanding Technical Contributions Award in 2024. He is a Fellow of the Singapore National Academy of Science.
\end{IEEEbiography}


\vfill

\end{document}